# Synergising Local Geo-Environmental Characteristics with Spatial Context for Enhancing Landslide Susceptibility Mapping

Yusen Cheng[1, 2], Lei Fan[1, *], Qinfeng Zhu[1, 3], Cheng Zhang[1], Yangyang Li[4, 5], Ron Mahabir[2]

[*] *Correspondence: Lei Fan (Lei.Fan@xjtlu.edu.cn)

[1] Department of Civil Engineering, Xi’an Jiaotong-Liverpool University, Suzhou, 215123, China

[2] Department of Geography and Planning, University of Liverpool, Liverpool, L69 3BX, UK

[3] Department of Computer Science, University of Liverpool, Liverpool, L69 3BX, UK

[4] Suzhou Industrial Park Monash Research Institute of Science and Technology, Monash University, No. 1 Huayun Road, SIP Suzhou, 215000, PR China

[5] Department of Civil Engineering, Monash University, 23 College Walk, Clayton, Victoria 3800, Australia

**Abstract**

Data-driven methods are widely used in landslide susceptibility mapping (LSM) because they can effectively model the complex relationships between landslides and geo-environmental conditions. Existing data-driven approaches generally follow two types of data representations. Pixel-based models focus solely on the geo-environmental characteristics of a specific landslide but neglect the influence of its surrounding environment. Patch-based models incorporate surrounding spatial context but may include pixels with weak or no spatial relevance to the target landslide location. To address this limitation, this study proposes a Local-Geo and Spatial Context Fusion (LGSCF) strategy, which synergises the geo-environmental characteristics of landslide points with their corresponding spatial context through a feature-wise modulation mechanism. We tested the LGSCF strategy by integrating it into several representative convolutional neural network (CNN) architectures, creating nine different LGSCF-based models. The primary study area covers approximately 2644 kmʇ across Jenai and Sinyi Townships in Nantou County, Taiwan, and the dataset comprises 5332 landslide samples and an equal number of non-landslide samples. The results show that LGSCF-based models consistently outperform their corresponding baselines, achieving F1-scores up to 87.09% and AUC values up to 0.9472. Furthermore, the susceptibility maps produced by LGSCF-based models show that known landslides are more accurately concentrated in "very high" susceptibility zones with fewer misclassifications. These findings demonstrate that our fusion strategy can significantly improve the accuracy of landslide susceptibility mapping.

# 1 Introduction

Landslides are downslope movements of rock, soil, or debris driven by gravity, commonly triggered by intense rainfall, earthquakes, volcanic activity, or human interventions [1], posing severe threats to both mountainous and densely populated urban regions [2,3]. In recent decades, the frequency of landslide disasters has shown a clear increase [4], closely associated with the combined effects of climate change and intensified human activities such as deforestation, slope modification, and unplanned urban expansion. These events often result in severe consequences, including loss of life, damage to infrastructure, and widespread socio-economic disruption [5,6]. Hence, accurate prediction of landslide-prone areas has become a crucial element of disaster prevention and mitigation, providing critical spatial insights for land-use planning and timely risk reduction.

In response to this need, landslide susceptibility mapping (LSM) has become one of the primary approaches. LSM refers to the delineation of spatial zones with different levels of susceptibility to landslides by analysing the relationship between historical landslide occurrences and landslide geo-environmental conditioning factors (LCFs) such as topography, geology, vegetation, and rainfall. The underlying assumption is that future landslides are more likely to occur under environmental conditions similar to those of past events. Over the years, LSM has transitioned from qualitative assessments based on expert knowledge and heuristic evaluations [7,8] to semi-quantitative approaches using statistical indices and weighted overlays [9,10]. More recently, with the advancement of data availability and computational techniques, quantitative methods have represented the prevailing practice [11ï 13]. Quantitative methods can be broadly divided into physically based and data-driven approaches. Physical models typically rely on geomechanical and hydrological principles of slope stability and are capable of providing detailed process-level insights, making them valuable in site-specific engineering practice. However, their application in regional-scale LSM is constrained by several factors, such as the difficulty of acquiring accurately measured soil and hydrological parameters [14,15], a focus on a single failure mechanism rather than diverse landslide types [16], and the high computational demand [17].

Given the ability and flexibility in handling diverse LCFs and in operating on landslide inventories from multiple sources, machine learning methods constitute the most widely applied category of data-driven approaches in LSM. Commonly used algorithms include logistic regression (LR) [18,19], support vector machines (SVM) [20,21], naµve Bayes [22,23], random forests (RF) [24,25], and k-nearest neighbours (KNN) [26,27]. In recent years, the application of machine learning in LSM has not only achieved consistently high prediction accuracy [28,29], but also provided insights into the relative importance of LCFs and the underlying mechanisms of slope failure through interpretability techniques [30,31]. Despite these strengths, machine learning models typically use raw or discretised LCFs as input features without extracting hierarchical representations to capture complex patterns. Moreover, when pixel units are adopted, classification is generally based only on the attributes of the landslide or non-landslide pixel itself, neglecting the spatial context of surrounding areas. These may lead to unstable predictions along slope boundaries and reduce the model's ability to capture terrain continuity and hydrological connectivity, thereby limiting the potential for further performance gains.

To address these limitations, Wang et al. [32] were the first to introduce deep learning into LSM, testing four different convolutional neural network (CNN) architectures (CNN1D, CNN2D, CNN3D, and LeNet-5) and reporting promising performance. Since then, CNN-based methods have been increasingly adopted in landslide studies and have frequently outperformed conventional machine learning approaches in terms of predictive accuracy [33–35]. From the perspective of input data representation, these methods can be broadly grouped into the following two categories.

The first category directly encodes the LCFs of landslide or non-landslide points without explicitly incorporating neighbourhood context. In this case, the LCFs of a sample are arranged as a one-dimensional array, with each feature placed sequentially along the same axis, or transformed into two-dimensional tensors after discretisation and reorganisation of categorical and continuous factors [36–38]. Although such CNN designs allow models to extract hierarchical representations and capture complex nonlinear relationships, they still rely solely on pixel-level attributes and therefore overlook the spatial context of surrounding conditions.

The second category expands around each landslide or non-landslide point to construct patch-based samples with dimensions $n \times n \times c$, where $n$ denotes the patch size, and c is the number of LCFs. This approach incorporates spatial context by including both the central pixel and its neighbouring environment [39–41]. Nevertheless, it inevitably introduces redundant information, as not all surrounding pixels are equally relevant to slope failure. Due to the nature of CNN operations, successive convolution and pooling layers can progressively aggregate neighbourhood signals and downsample the feature maps, so the contribution of the central pixel, which directly corresponds to the class label, may become blurred or diluted. Furthermore, although the final fully connected (FC) layers can learn nonlinear combinations and implicitly weight features, they lack an explicit mechanism to prioritise critical information or to condition the spatial representation accordingly. As a result, the model may be overly influenced by irrelevant background features while underutilising the most informative cues. These limitations have motivated two related directions in recent LSM research: applying architectures that can model broader spatial dependencies and designing fusion mechanisms that can integrate complementary representations more effectively.

Recent transformer and graph-based approaches primarily reflect the first direction. Vision Transformers (ViTs) treat image patches as tokens and use self-attention to capture long-range interactions, although their performance may be constrained by the limited training samples typical of LSM [42]. Hybrid CNNï Transformer models combine the extraction of terrain patterns by CNNs with transformer-based global-context modelling [43,44]. Graph neural networks (GNNs), by contrast, represent slope units or environmentally related locations as nodes and encode their spatial or environmental relationships as edges, allowing information to propagate between adjacent or similar terrain units [45ï 47]. These approaches extend the spatial modelling capability of conventional CNNs, but they mainly focus on feature extraction and dependency modelling rather than explicitly coordinating the roles of local environmental attributes and surrounding geospatial context.

The second direction concerns how complementary representations are fused. In remote sensing, heterogeneous information can be combined at the input, intermediate-feature, or decision level [48]. At the feature level, fusion strategies range from direct concatenation or summation to learned feature recalibration. Squeeze-and-excitation (SE) and convolutional attention modules learn channel-wise and/or spatial weights to emphasise informative responses [49,48], whereas transformer-based approaches use self-attention or cross-attention to model long-range and cross-source dependencies [50]. However, these methods are not specifically designed to address a particular challenge in point-centred, patch-based LSM: the central pixel represents the location to be classified, whereas the surrounding patch provides contextual information whose relevance may vary according to the local geo-environmental conditions. Thus, the two inputs are not interchangeable modalities, but two representations of the same LCF data at different spatial scopes.

To address the issues, this study proposes a Local-Geo and Spatial Context Fusion (LGSCF) framework that synergises pixel-level geo-environmental characteristics with their surrounding spatial context. Specifically, the local-geo branch encodes the geo-environmental characteristics of the central pixel, corresponding to the location of a landslide or non-landslide point, while the spatial-context branch captures spatial patterns from the entire image patch. The two branches are then integrated through a feature-wise linear modulation mechanism. This framework provides a flexible design that allows different CNN architectures to be paired as the two branches. In this way, the model leverages informative neighbourhood context while reducing redundancy and prioritises features most predictive of the class label.

## 2 Study area and data

### 2.1 Study area description

Nantou County is located in the central part of Taiwan and forms a key section of the island's mountain range (Fig. 1a). The topography varies greatly, with elevations ranging from about 20 m in the western lowlands, which are mainly occupied by built-up areas, to approximately 3800 m in the eastern mountains characterised by steep slopes and complex geological formations. According to the Köppen climate classification [51], Nantou falls within the humid subtropical climate zone (Cfa), strongly influenced by the East Asian monsoon. Precipitation varies considerably due to the blocking effect of the mountainous terrain on moisture-laden air masses, with annual averages of around 1750 mm in the western plains and more than 2800 mm in the eastern mountains [52]. Most rainfall occurs between April and September during the monsoon season. The geological and climatic conditions make Nantou highly susceptible to landslides, making it an important and representative site for landslide-related research [53,54]. In this study, Jenai and Sinyi Townships, located in the eastern part of Nantou County and covering a combined area of about 2,644 km$^2$, were selected as the study area (Fig. 1b). To further examine whether the effectiveness of the proposed LGSCF framework can be reproduced beyond the primary study area, Hualien County in eastern Taiwan was additionally selected for cross-regional validation. Despite their geographical proximity, Nantou and Hualien differ in their environmental settings, with Nantou characterised by an inland mountainous landscape and Hualien by an eastern coastal setting, leading to differences in terrain and rainfall regimes.

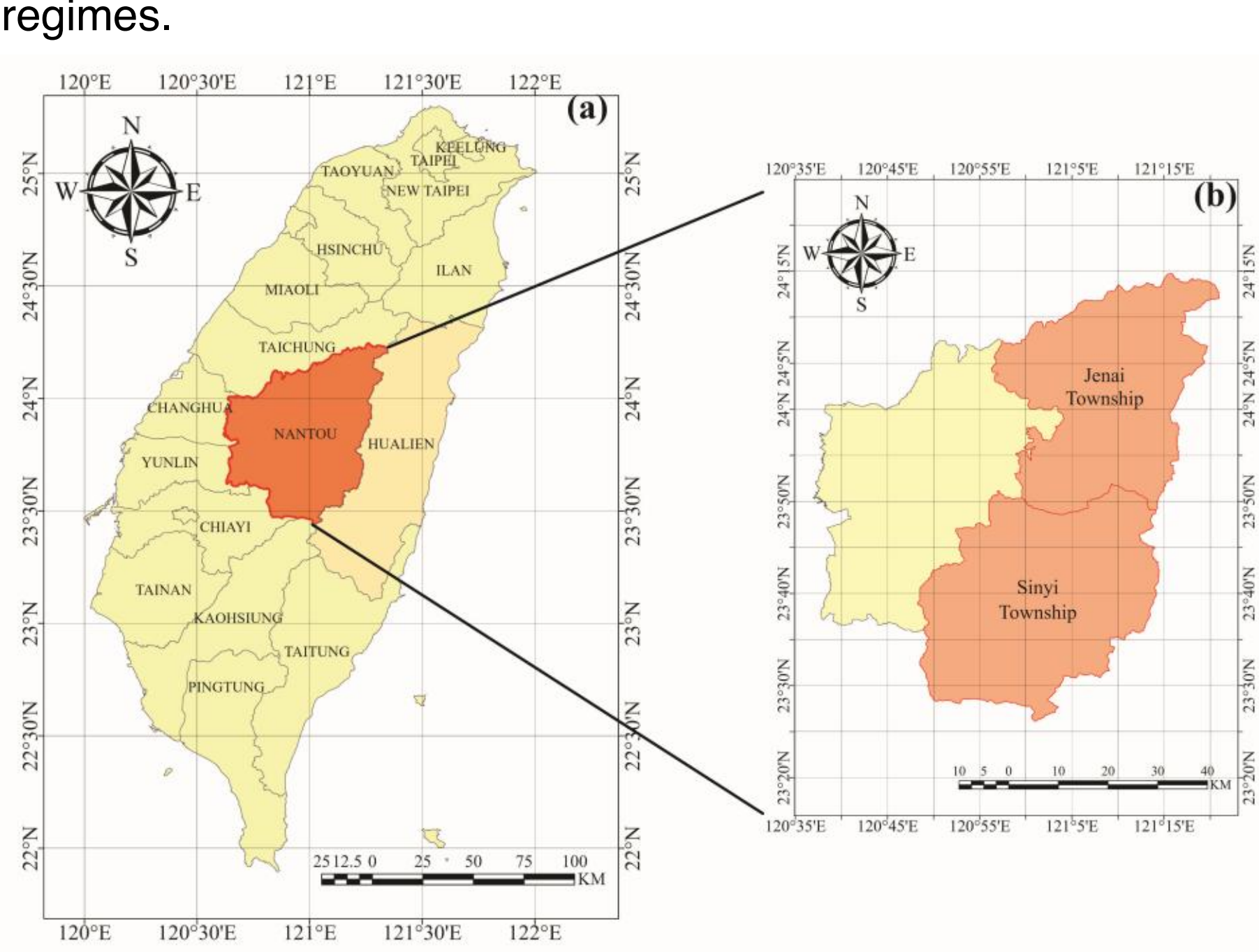

Fig. 1. Location of (a) Nantou County and Hualien County in Taiwan and (b) the primary study area

comprising Jenai and Sinyi Townships in eastern Nantou County.

## 2.2 Landslide inventory

The accuracy of landslide points is critical for reliable landslide susceptibility analysis [55]. In this study, the landslide inventory was derived from the 2022 Annual Landslide Inventory Map of Taiwan [56]. The map was produced through visual interpretation of SPOT satellite imagery with a minimum mapping unit of 0.1 hectares. Crown areas are commonly used as representative points in landslide susceptibility analysis [57]. In this study, 5332 landslide points were extracted within the study area (Fig. 2).

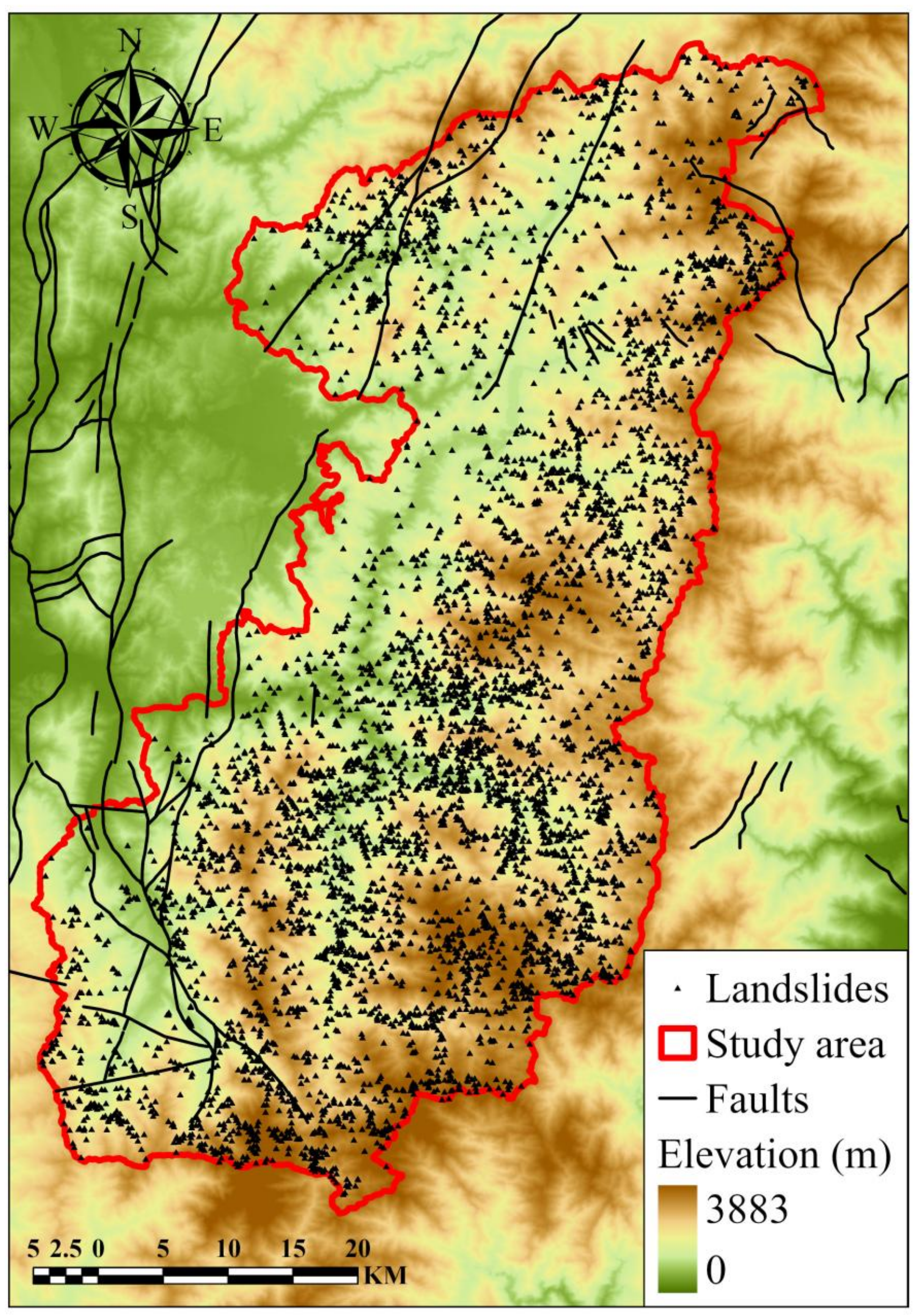

**Fig. 2**. Spatial distribution of the 5332 landslide inventory points within the study area.

## 2.3 Landslide geo-environmental conditioning factors (LCFs)

Data-driven LSM estimates the likelihood of landslide occurrence by learning the relationships between observed landslides and their geo-environmental settings. The selection of appropriate LCFs is therefore a critical step. Although numerous LCFs have been used in previous studies, no universally accepted selection method exists because their relevance depends on the landslide type, spatial scale, regional setting, and data availability [58ï 61]. In this study, fifteen LCFs were selected to represent five groups of landslide controls: topographic, geological, hydrological, surface environmental, and human-related factors. The selection was guided by their established use in regional LSM, their physical relevance to landslide occurrence, and the availability of datasets with consistent coverage across Nantou County. These factors are particularly relevant to the steep terrain, complex geological setting, monsoon-dominated rainfall, variable land cover, and road development that characterise Nantou County [52ï 54].

Other potentially relevant factors were considered but not included in the present analysis. Soil type was excluded because a region-wide dataset with spatial resolution and attribute consistency comparable to other inputs was unavailable. Lithology and LULC were therefore retained to characterise surface conditions, although they cannot fully substitute for detailed soil information. Landform classes were not included as separate inputs because they are commonly derived from DEM-based terrain attributes already represented by elevation, slope, aspect, curvature, TRI, TWI, and SPI. Plane and profile curvature were included among the fifteen LCFs, whereas additional curvature derivatives were omitted to avoid introducing closely related terrain variables. Seismicity was also excluded because this study addresses general landslide susceptibility rather than an event-specific coseismic scenario. A meaningful seismic predictor, such as peak ground acceleration, would require spatial ground-motion data temporally matched to the landslide inventory [62ï 64].

All LCFs were processed into raster format and resampled to a uniform spatial resolution of 30 m Ĭ 30 m to ensure consistency in subsequent modelling. The details and spatial distributions of these LCFs are presented in Table 1 and Fig. 3.

**Table 1**. Data sources for LCFs used in this study

| Data | Sources | LCFs |
|---|---|---|
| DEM | ASTER Global Digital Elevation Model (GDEM) V003 (30m) | elevation, slope, aspect, plane curvature, profile curvature, stream power index, TWI, and topographic roughness index. |
| lithology | Derived from the ñTaiwan National | Lithology |

| fault lines | Land Surveying and Mapping Information Web Map Serviceòand manually vectorised | Distance to faults |
|---|---|---|
| NDVI | Chinese Academy of Science Discipline Data Centre for Ecosystem (30m) [65] | NDVI |
| LULC | The global 2000-2020 LULC dataset (30m) [66] | LULC |
| rainfall | CMORPH Climate Data Record (CDR), resampled to 30m | rainfall |
| road networks | OpenStreetMap | Distance to roads |
| river networks | | Distance to rivers |

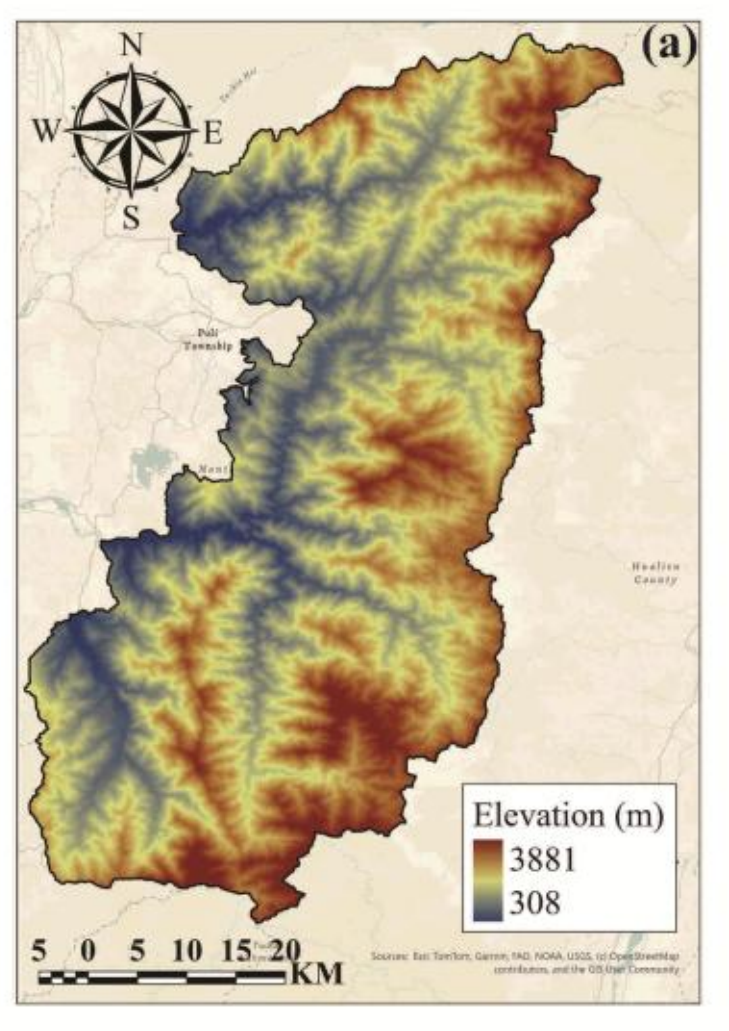

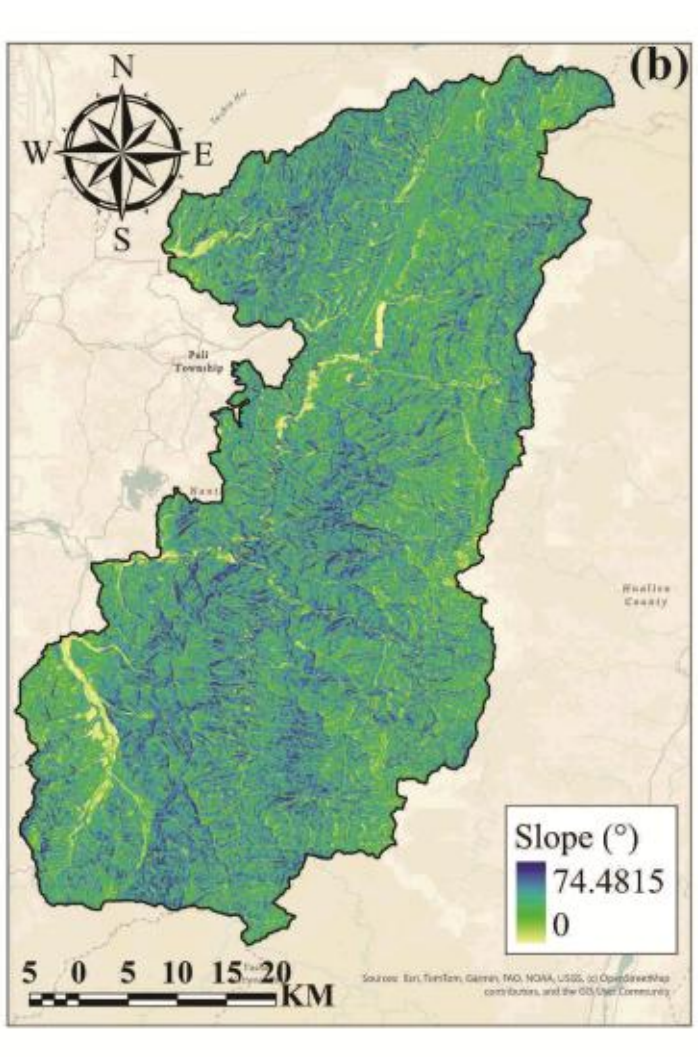

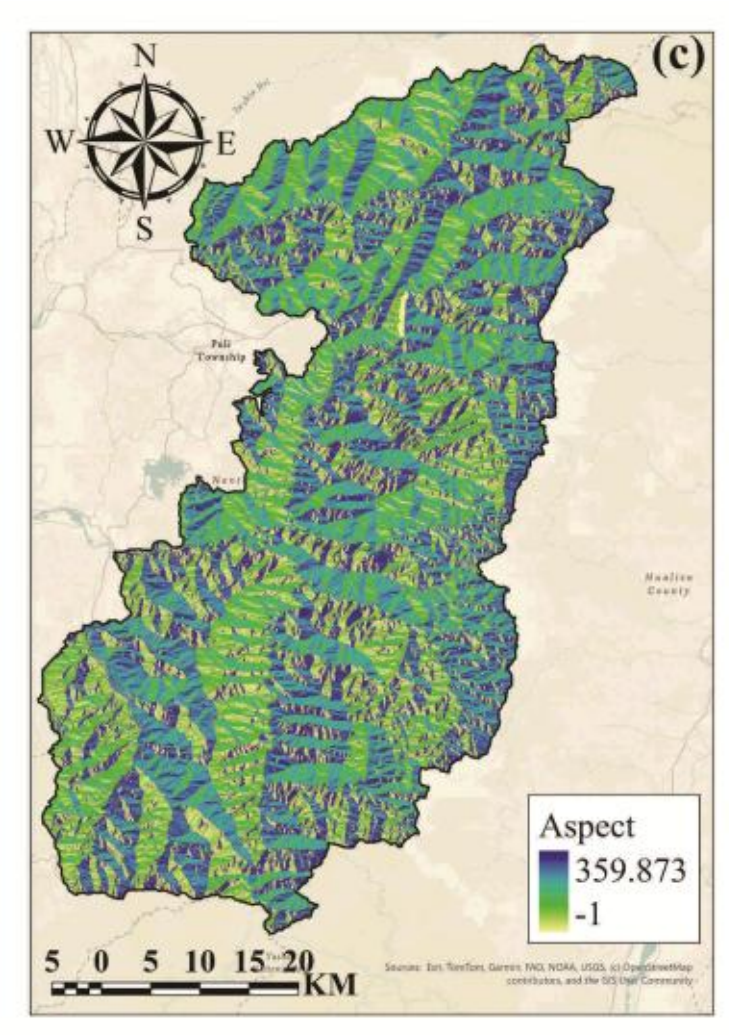

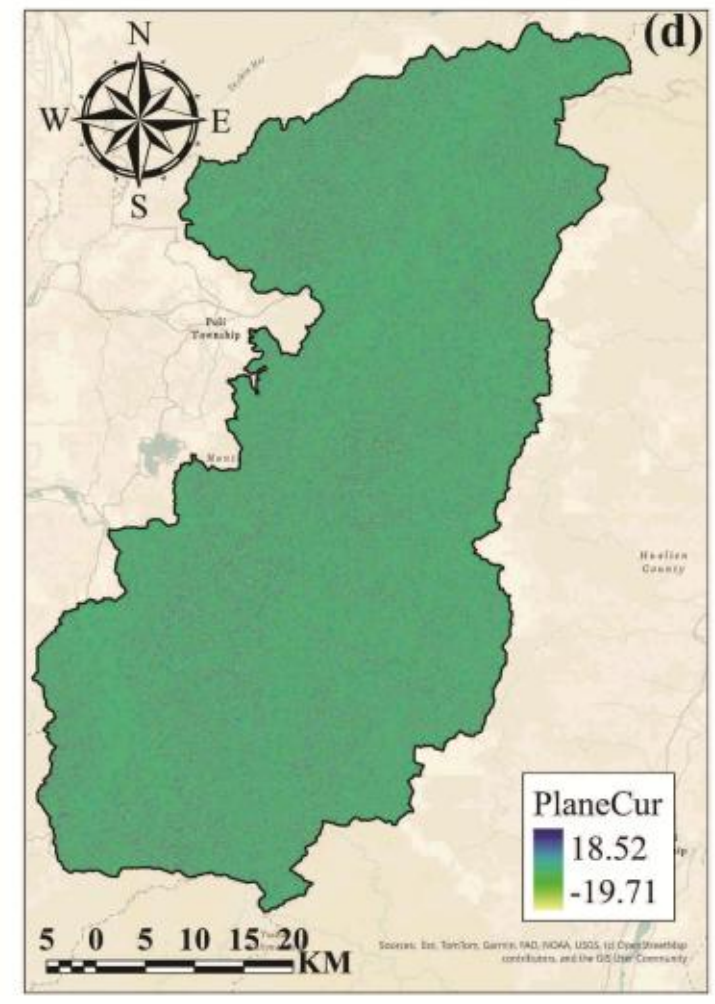

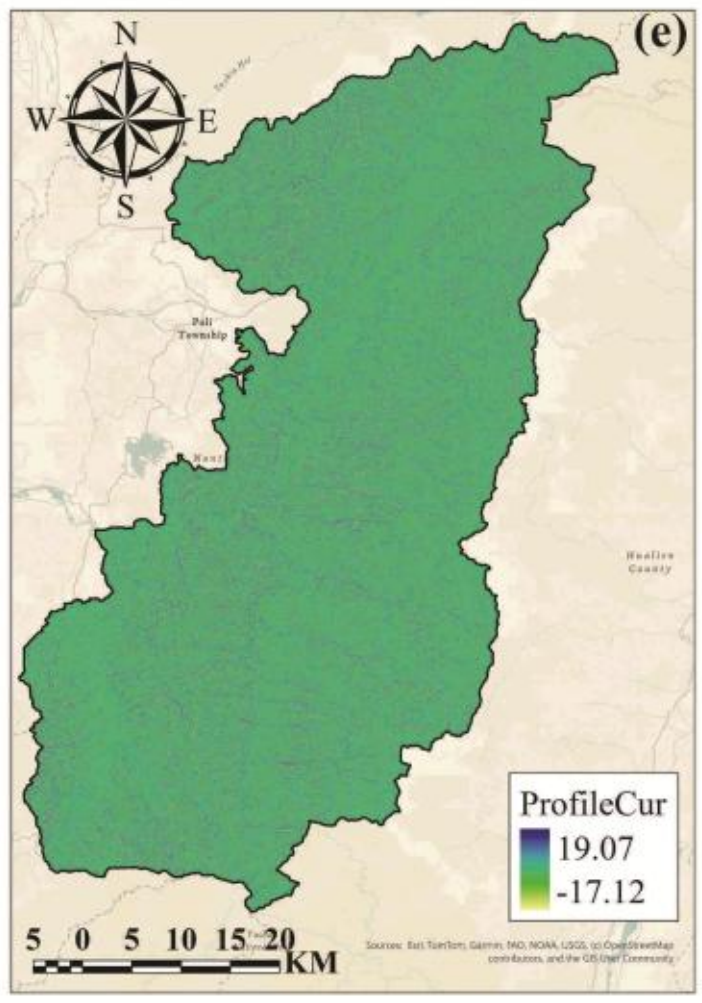

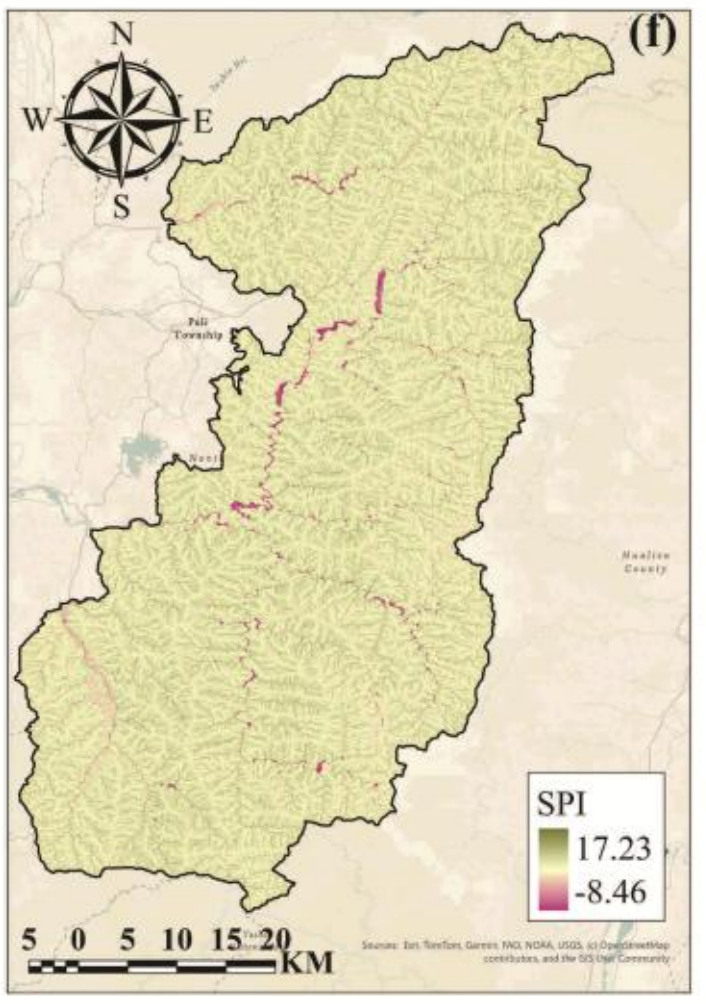

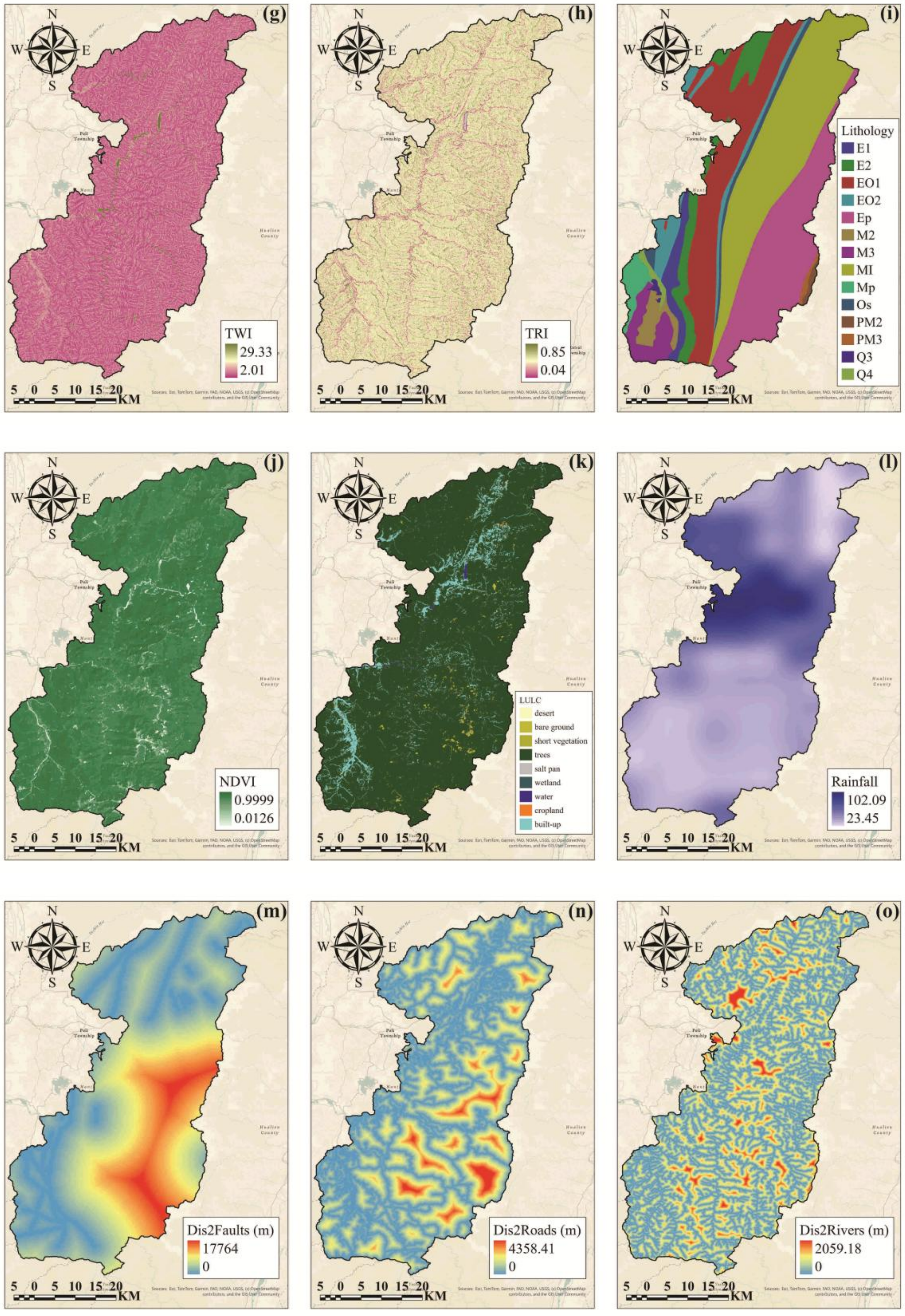

**Fig. 3**. Spatial distributions of the 15 LCFs used in this study: (a) elevation; (b) slope; (c) aspect; (d) plane curvature; (e) profile curvature; (f) SPI; (g) TWI; (h) TRI; (i) lithology; (j) NDVI; (k) LULC; (l) rainfall; (m) distance to faults; (n) distance to roads; (o) distance to rivers.

# 3 Methodology

## 3.1 LCFs multicollinearity diagnostics

Reliable landslide susceptibility modelling requires that the input LCFs do not exhibit strong linear dependency, as strongly correlated LCFs may obscure the individual contribution of each factor and affect model stability. To address this, we employed two commonly used multicollinearity diagnostic measures: tolerance (TOL) and the variance inflation factor (VIF), as shown in Formula (1).

$$VIF = \frac{1}{1 - R^2} = \frac{1}{TOL} \tag{1}$$

where $R^2$ is the coefficient of determination obtained by regressing one factor against all others, and *VIF* is the reciprocal of *TOL,* which quantifies how much the variance of a regression coefficient is inflated due to correlation among factors.

In general, TOL values lower than 0.1 and VIF values greater than 10 are regarded as the existence of multicollinearity.

## 3.2 Dataset construction

Data-driven landslide susceptibility modelling is essentially a binary classification task, which requires both landslide points (positive samples) and non-landslide points (negative samples). A variety of strategies for selecting non-landslide points have been proposed and applied in previous studies [67ï 69]. In this study, the frequency ratio (FR) model was adopted to select non-landslide samples [70ï 72]. The FR-based sampling strategy was intended to reduce the risk of treating unrecorded or potentially unstable locations as non-landslide samples, because the absence of an inventory record does not necessarily confirm slope stability.

The FR method is a statistical approach that measures the relationship between the spatial distribution of landslides and the corresponding classes of LCFs. FR values were calculated using Formula (2) and then normalised to relative frequency (RF), as shown in Formula (3). Based on these results, the prediction rate (PR) was derived according to Formula (4). To minimise the likelihood of mislabelling unstable areas as negative samples, the product of each factorȇs PR and the RF of its classes was calculated, and non-landslide samples were randomly selected from the very low susceptibility zones to match the number of landslide samples.

$$FR_{ij} = \frac{LA_{ij} / LA_t}{A_{ij} / A_t} \tag{2}$$

$$RF_{ij} = \frac{FR_{ij}}{\sum_{k=1}^{m_i} FR_{ik}} \quad (3)$$

$$PR_i = \frac{(RF_{i,max} - RF_{i,min})}{min_{r=1,2,\ldots,n}(RF_{r,max} - RF_{r,min})} \quad (4)$$

where $i$ denotes the $i$-th LCF and $j$ denotes its $j$-th category. $LA_{ij}$ and $A_{ij}$ represent the landslide area and the total area within category $j$ of LCF $i$, respectively, while $LA_t$ and $A_t$ are the total landslide area and the total area of the study area. $FR_{ij}$ is the frequency ratio of category $j$ within LCF $i$. A value greater than 1 indicates that landslides are proportionally more concentrated in that category than in the study area as a whole. In Equation (3), $m_i$ is the number of categories within LCF $i$, and $RF_{ij}$ is the normalised relative frequency, for which the values of all categories within the same LCF sum to 1. In Equation (4), $RF_{i,\max}$ and $RF_{i,\min}$ are the maximum and minimum $RF$ values among the categories of LCF $i$, respectively; $n$ is the total number of LCFs; and $r$ indexes the LCFs. $PR_i$ is the prediction rate of LCF $i$, obtained by normalising its RF range against the smallest RF range among all LCFs.

To ensure class balance, an equal number of non-landslide points (5332) were selected to match the landslide points. The selected landslide and non-landslide points were organised into image patches to construct the dataset. A previous study has shown that the side length of image patches can influence modelling performance, and patch sizes that approximate the average landslide size within the study area tend to yield better results [73]. In our study, the average landslide area is approximately 16333.7 mȴ, corresponding to a side length of about 127.8 m. Given the resolution of 30 m, an 11Ĭ 11 image sample was generated by extending five pixels outward in all directions from each selected point (either landslide or non-landslide). This sample size was chosen to ensure adequate coverage of the typical landslide extent while maintaining a manageable input size for the following modelling procedure. Combined with the 15 LCFs, each image patch sample thus has a structure of 11Ĭ 11Ĭ 15. The dataset was then divided into training and validation sets in a 7:3 ratio.

## 3.3 Modelling description

To address the limitations of previous methods stated in Section 1, this study proposes the LGSCF strategy for landslide susceptibility mapping, based on a dual-branch CNN design (Fig. 4). The input sample is denoted as $X \in \mathbb{R}^{11\times11\times15}$. The spatial-context branch takes the full patch $X$ as input and is designed to capture both the landslide or non-landslide characteristics and the surrounding contextual patterns. After convolutional processing, the spatial-context branch outputs a one-dimensional feature vector $s \in \mathbb{R}^k$. In parallel, the local-geo branch encodes a 15-dimensional vector consisting of the LCF values of the centre pixel, which corresponds to either a landslide or non-landslide location, denoted as $X_c \in \mathbb{R}^{15}$. After 1D convolution, the local-geo branch yields a feature representation $c \in \mathbb{R}^{D_c}$.

Regarding feature fusion, a straightforward approach would be concatenation. However, this design faces two potential issues. First, as the spatial-context branch inherently includes the centre pixel information, direct concatenation with the local-geo branch may introduce feature redundancy. Second, the two branches process inputs of different dimensionality and semantic scope (patch-level vs pixel-level), so direct concatenation could result in inconsistency in physical meaning. Therefore, in this study, we draw inspiration from Feature-wise Linear Modulation (FiLM) [74,75]. The core idea of FiLM is to condition one feature representation on another by generating feature-wise scaling and shifting parameters. In this study, specifically, the local-geo representation $c$ is transformed into modulation parameters $\gamma, \beta \in \mathbb{R}^k$ through a two-layer multilayer perceptron (MLP) (Formula 5), which are then used to perform feature-wise affine modulation of the spatial representation $s$ (Formula 6).

$$[\gamma, \beta] = g(c; \theta_g) \in \mathbb{R}^{2k} \quad (5)$$

where $\theta_g$ denotes the learnable parameters of the MLP.

$$f = \gamma \odot s + \beta \quad (6)$$

where $\odot$ denotes feature-wise multiplication. The fused vector $f$ is subsequently fed into fully connected layers for final classification.

Conceptually, LGSCF is an intermediate feature-level fusion strategy rather than a generic multimodal learning paradigm. It differs from conventional attention-based fusion, in which channel, spatial, or token attention generally derives weights over feature channels, spatial locations, or token-to-token interactions. LGSCF does not construct an attention map or calculate similarity between feature tokens. Instead, it performs cross-branch affine modulation, in which the local-geo representation generates both multiplicative and additive parameters for the spatial-context features. The scaling operation strengthens or suppresses individual contextual responses, while the additive shift allows the conditioned representation to move beyond simple multiplicative reweighting. This mechanism is particularly relevant to LSM because the class label corresponds to a specific central location, whereas the surrounding patch may contain informative or unrelated terrain patterns. The relevance of this contextual information depends on the geo-environmental conditions at the target location. For instance, similar neighbourhood patterns may have different implications under different topographic, geological, or hydrological settings. LGSCF therefore uses the central-location characteristics to guide how the surrounding context is interpreted, providing a task-specific form of directional conditioning.

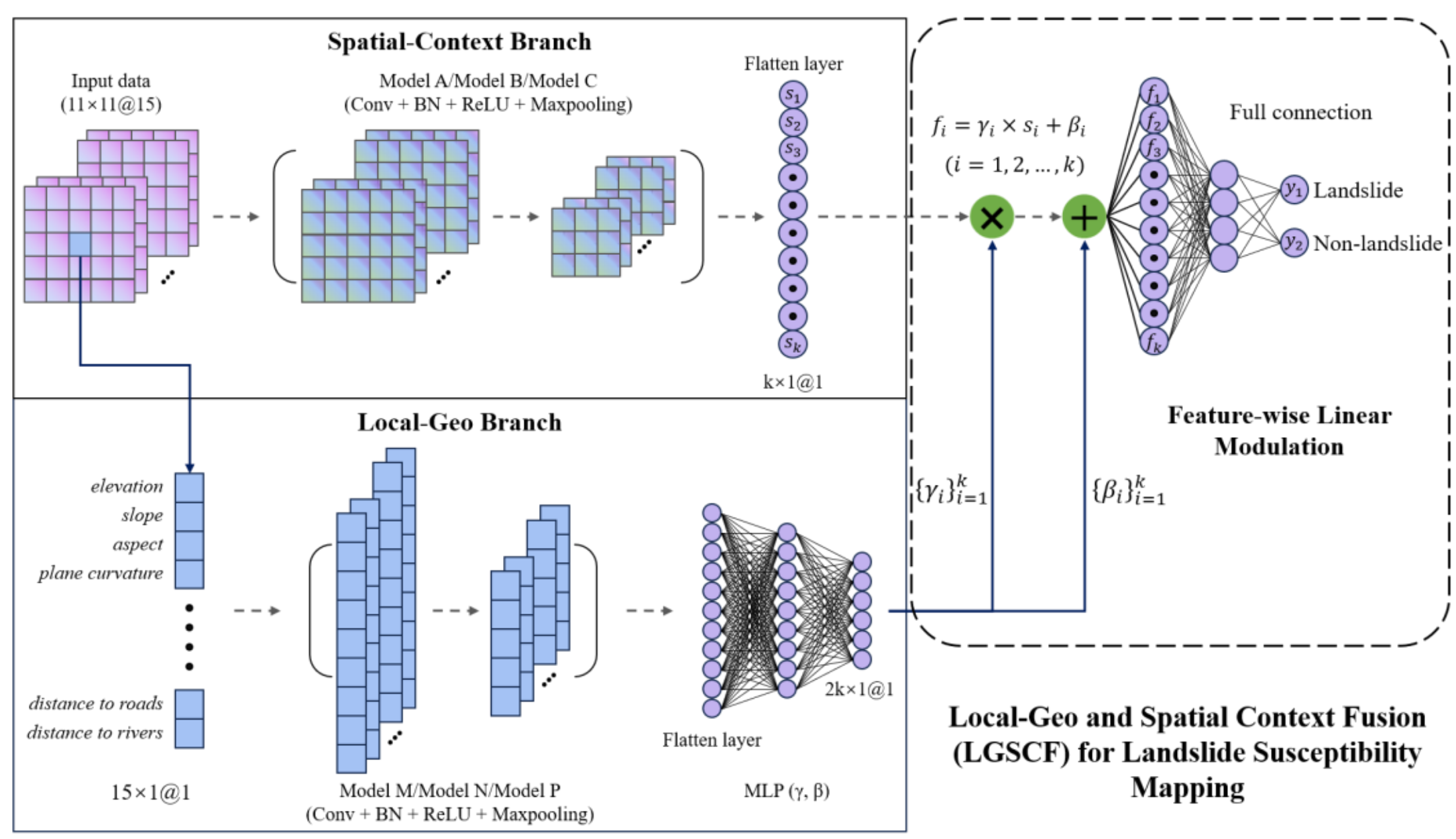

**Fig. 4**. Architecture of the proposed LGSCF strategy. Models A, B, and C represent the spatial-context branch, while Models M, N, and P correspond to the local-geo branch. Further details are provided in Table 3.

To evaluate the effectiveness of the proposed LGSCF strategy, we selected six representative baseline CNN architectures, including three spatial-context branch models (denoted as Models A, B, and C) and three local-geo branch models (denoted as Models M, N, and P). Details of these baseline models are summarised in Table 3. By combining each model for the spatial-context branch with that for the local-geo branch in turn, nine LGSCF-based models were constructed for comparison, as listed in Table 4. Combinations within the same branch type (for example, two spatial-context branches or two local-geo branches) were not considered, as they cannot provide complementary information and thus hold little practical meaning. In all cases, the fully connected layers of the baseline models were removed, and outputs up to the flatten layer were retained to enable feature-level fusion.

**Table 3**. Baseline models used in this study.

| Model ID | References | Main architectural characteristics | Branch type |
|---|---|---|---|
| Model A | [32] | 3D convolution-based feature extraction. | Spatial-context branch |
| Model B | [76] | CNN with sequential channel and spatial attention (CBAM). | |
| Model C | [77] | CNN component adapted from a physically-based probabilistic model with CNN (PPM-CNN) framework. | |
| Model M | [32] | 1D convolution-based feature extraction. | Local-geo branch |
| Model N | [37] | 1D CNN with hierarchical convolution and pooling. | |

| Model P | [38] | CNN component adapted from the CNN and deep forest (CNN-DF) framework. |
|---|---|---|

**Table 4**. List of LGSCF-based models and their corresponding baselines.

| **Model ID** | **Model M** | **Model N** | **Model P** |
|---|---|---|---|
| **Model A** | Model A+M | Model A+N | Model A+P |
| **Model B** | Model B+M | Model B+N | Model B+P |
| **Model C** | Model C+M | Model C+N | Model C+P |

**Notes**: Models A, B, and C represent the spatial-context branch, while Models M, N, and P correspond to the local-geo branch, as detailed in Table 3. Combinations within the same branch type are not applicable to the proposed LGSCF strategy.

Hyperparameter optimisation was conducted independently for each baseline and LGSCF-based model using an exhaustive grid search over batch size and initial learning rate (Table 5). Two batch sizes (32 and 64) and five initial learning rates ($10^{-5}$, $10^{-4}$, $10^{-3}$, $10^{-2}$, $10^{-1}$) were considered, resulting in ten candidate configurations for each model. To reduce the influence of random weight initialisation and data shuffling, each configuration was independently trained eight times using random seeds from 0 to 7. All candidate configurations used the Adam optimiser, cross-entropy loss, and a polynomial learning-rate scheduler with a decay power of 0.9. Training was limited to 200 epochs, with early stopping applied when the validation loss failed to improve by more than $1 \times 10^{-4}$ for ten consecutive epochs. For each model, the candidate configurations were ranked according to their mean validation AUC across the eight runs, and the configuration with the highest mean AUC was selected for subsequent evaluation. Using the mean performance rather than the best individual run reduced the likelihood of selecting a configuration based on a favourable random initialisation. All other training settings were kept unchanged to ensure a consistent comparison among models.

**Table 5**. Training configurations of baseline models and LGSCF-based models.

| **Settings** | **Values** |
|---|---|
| Batch sizes | 32, 64 |
| Initial learning rates | $10^{-5}$, $10^{-4}$, $10^{-3}$, $10^{-2}$, $10^{-1}$ |
| Learning rate scheduler | Polynomial decay (PolyLR), with decay power $p = 0.9$ |
| Optimizer | Adam |
| Loss function | Cross-entropy |
| Number of epochs | 200 |
| Early stopping | Monitored on validation loss, with tolerance 1Ĭ 10ï4 and patience of 10 epochs. |
| Random seeds | For each setting, training was repeated with 8 seeds (0ï 7) to reduce variance and report stable statistics. |

## 3.4 Accuracy assessment strategy

Accurate evaluation is essential to measure the reliability and performance of data-driven models in LSM. In this study, all evaluations were computed on the validation set. Several complementary metrics commonly used in classification tasks are employed. These metrics are derived from the confusion matrix (Fig. 5), which consists of true positives (TP, correctly identified landslides), true negatives (TN, correctly identified non-landslides), false positives (FP, non-landslides incorrectly identified as landslides), and false negatives (FN, landslides incorrectly identified as non-landslides). Based on these quantities, overall accuracy is defined as the proportion of correctly classified samples among all samples. Precision measures the fraction of correctly predicted landslides among all predicted landslides, while recall measures the fraction of actual landslides that are correctly detected. The F1 score, defined as the harmonic mean of precision and recall, accounts for both FP and FN and provides an overall indicator of classification effectiveness.

| Actual \ Predicted | Landslide | Non-landslide |
| --- | --- | --- |
| Landslide | True Positive (TP) | False Negative (FN) |
| Non-landslide | False Positive (FP) | True Negative (TN) |

$$Accuracy = \frac{TP + TN}{TP + TN + FP + FN}$$

$$Precision = \frac{TP}{TP + FP}$$

$$Recall/Sensitivity = \frac{TP}{TP + FN}$$

$$F1 - score = \frac{2 \times Precision \times Recall}{Precision + Recall}$$

**Fig. 5**. Binary confusion matrix and equations used to calculate accuracy, precision, recall (sensitivity), and F1-score.

In addition to the confusion matrix, the receiver operating characteristic (ROC) curve is also a commonly used and reliable tool for evaluating model performance in LSM [78ï 80]. The ROC curve plots the true positive rate (TPR) against the false positive rate (FPR) at different decision thresholds, providing an overall view of classification ability. The area under the ROC curve (AUC) is a quantitative measure derived from the ROC curve. An AUC value closer to 1 indicates stronger discriminative power, while a value near 0.5 suggests performance comparable to random guessing.

Beyond classification metrics, Root Mean Squared Error (RMSE) and Mean Absolute Error (MAE) were further used to assess the numerical consistency between predicted susceptibility values and reference labels. RMSE penalises larger errors more heavily (Formula 7), while MAE reflects the average magnitude of errors. Lower values indicate more accurate susceptibility predictions (Formula 8).

$$RMSE = \sqrt{\frac{\sum_{i=1}^{M}(y_i - y_i')^2}{M}} \quad (7)$$

$$MAE = \frac{\sum_{i=1}^{M}|y_i - y_i'|}{M} \quad (8)$$

where $M$ is the total number of samples used for the evaluation. $y_i$ denotes the actual value for the $i$-th sample, while $y_i'$ denotes the predicted value for the $i$-th sample.

## 3.5 Model interpretation using SHAP

SHapley Additive exPlanations (SHAP) was used to interpret the contribution of each LCF to model predictions. SHAP is an additive feature-attribution method derived from the Shapley value in cooperative game theory [1,2]. It treats the input features as players and measures the contribution of each feature by considering its marginal effect across all possible feature combinations. For a sample $x$, the SHAP value of feature $i$ is calculated as:

$$\phi_i(x) = \sum_{S \subseteq F \setminus \{i\}} \frac{|S|!(M - |S| - 1)!}{M!}[v_x(S \cup \{i\}) - v_x(S)] \quad (9)$$

where $F$ is the complete set of $M$ input features, $S$ is a subset of features not containing feature $i$, and $v_x(S)$ denotes the expected model output when only the features in $S$ are known. The weighting term accounts for all possible orders in which feature $i$ can be added to a feature combination. The model prediction can then be expressed as:

$$f(x) = \phi_0 + \sum_{i=1}^{M} \phi_i(x) \quad (10)$$

where $\phi_0 = E[f(X)]$ is the expected model output and $\phi_i(x)$ represents the contribution of feature $i$ to the prediction.

A positive SHAP value indicates that the feature increases the predicted landslide susceptibility relative to the expected output, whereas a negative value indicates the opposite. Local SHAP values explain individual predictions, while global feature importance was evaluated by averaging the absolute SHAP values of each feature across all samples.

# 4 Results

## 4.1 Results of LCFs multicollinearity diagnostics

As shown in Fig. 6, the TOL values of the 15 selected LCFs ranged from 0.487 to 0.953, with the highest value observed for aspect (0.953) and the lowest for plane curvature (0.487). Correspondingly, the VIF values ranged from 1.049 to 2.053, with aspect exhibiting the lowest VIF (1.049) and plane curvature the highest (2.053). The VIF values were all below the commonly used threshold of 10, and all TOL values exceeded 0.1. The results indicated that the multicollinearity among the LCFs was insignificant, and all LCFs were therefore suitable for subsequent susceptibility analysis.

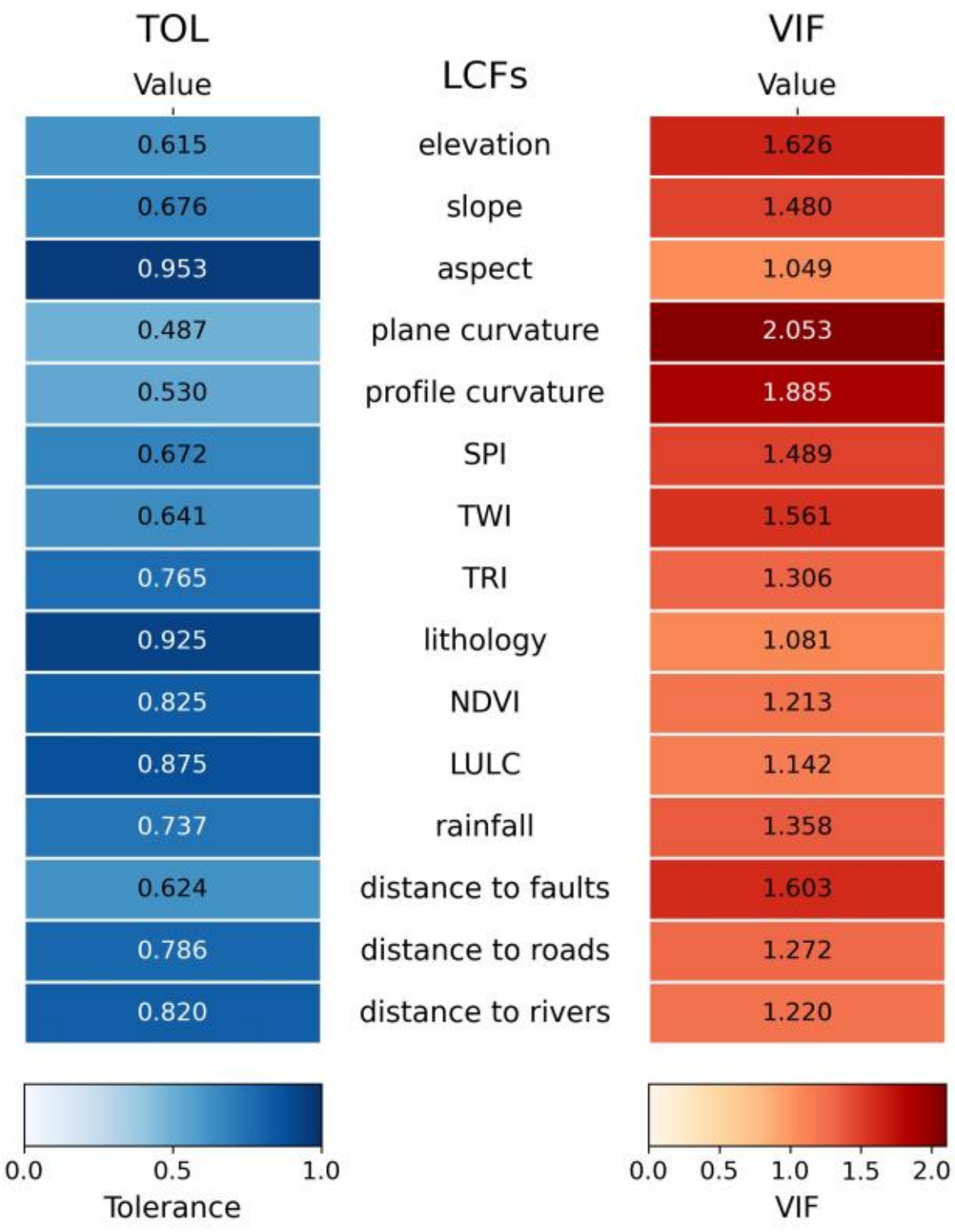

**Fig. 6**. Multicollinearity diagnostics (TOL and VIF) for the LCFs used in this study.

## 4.2 Configurations of best-performing models

As illustrated in Table 5, various batch sizes and learning rates were explored during model training to ensure a fair comparison across different architectures. Each configuration was run multiple times using random seeds. The mean AUC values were calculated and used to rank the configurations. Fig. 7 summarises the optimal settings corresponding to the best predictive performance for each model. The results show that a batch size of 32 consistently yielded competitive performance across most model variants, while a larger batch size of 64 was more frequently selected by the LGSCF-based models by enhancing gradient stability. Regarding the learning rate, most models achieved their best performance with values of $10^{-4}$ or $10^{-3}$, while only a few models selected $10^{-2}$, and none chose $10^{-5}$ or $10^{-1}$.

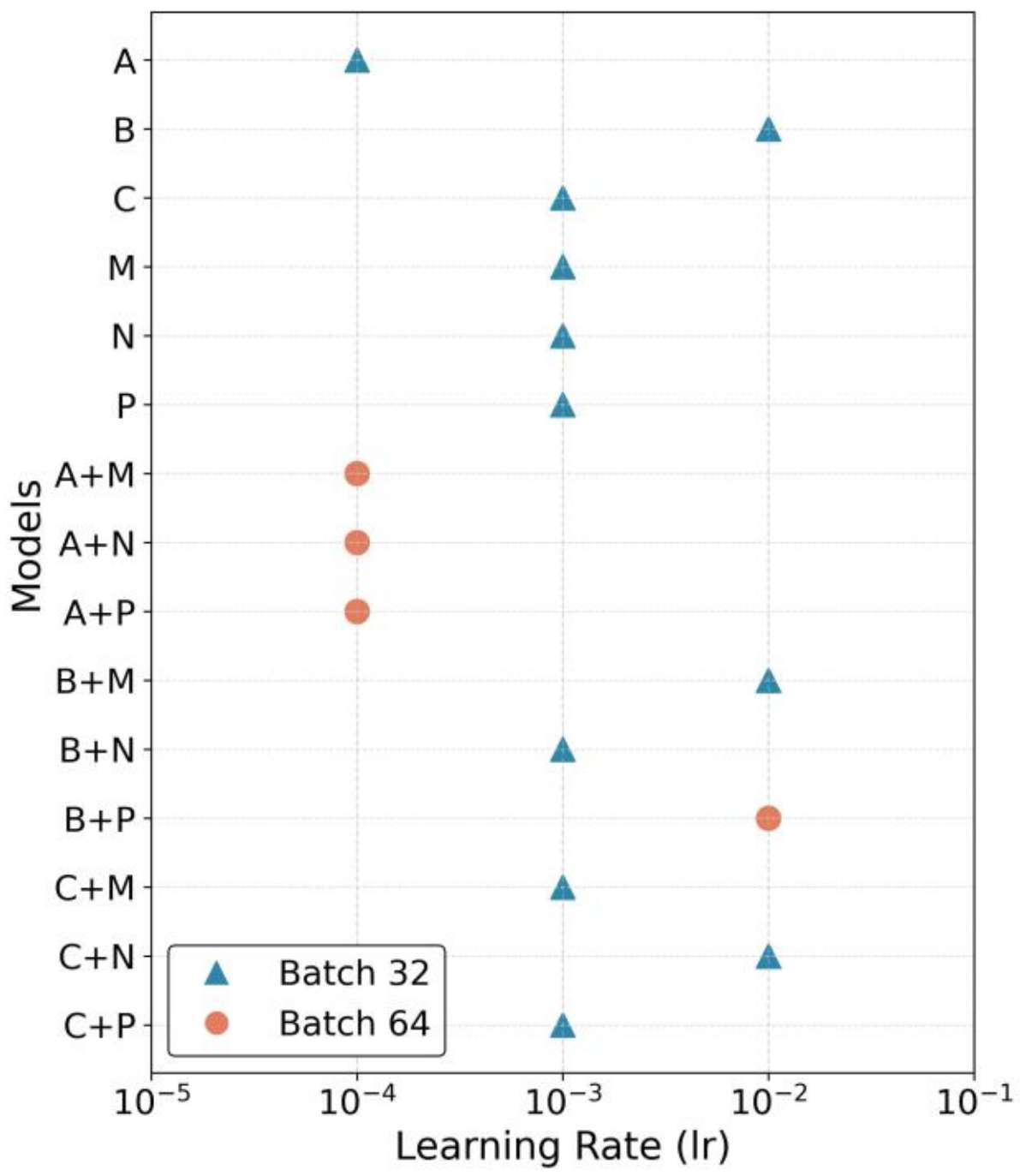

**Fig. 7**. Batch size and initial learning rate selected for each baseline and LGSCF-based model based on the highest mean validation AUC across eight repeated runs.

## 4.3 Susceptibility map of the study area.

To visualise the predictive results, we used the best-performing run to generate the landslide susceptibility map for each model. The outputs were classified into five susceptibility levels: "very low", "low", "moderate", "high", and "very high", using the geometric interval method [81,82]. Fig. 8 presents the susceptibility maps of Model A+M together with its two baselines, Models A and M, while the maps from other models are provided in Appendix A.

Overall, the three maps display broadly consistent spatial patterns: "very high" susceptibility zones were concentrated in the southeast, followed by the northeast, whereas the western region was dominated by "low" and "very low" susceptibility levels. This distribution aligns well with the topography and the actual landslide distribution, as the mountainous east is more prone to slope failures than the flatter west. Looking more closely, however, some differences can be noted in how the models delineated susceptibility levels. Within the "high" susceptibility areas, Model A+M still identified patches of "very low" susceptibility, while Models A and M tended to assign these zones as "moderate". Conversely, in "very low" susceptibility regions, Model A+M assigned more "moderate" zones, reflecting a more conservative classification behaviour.

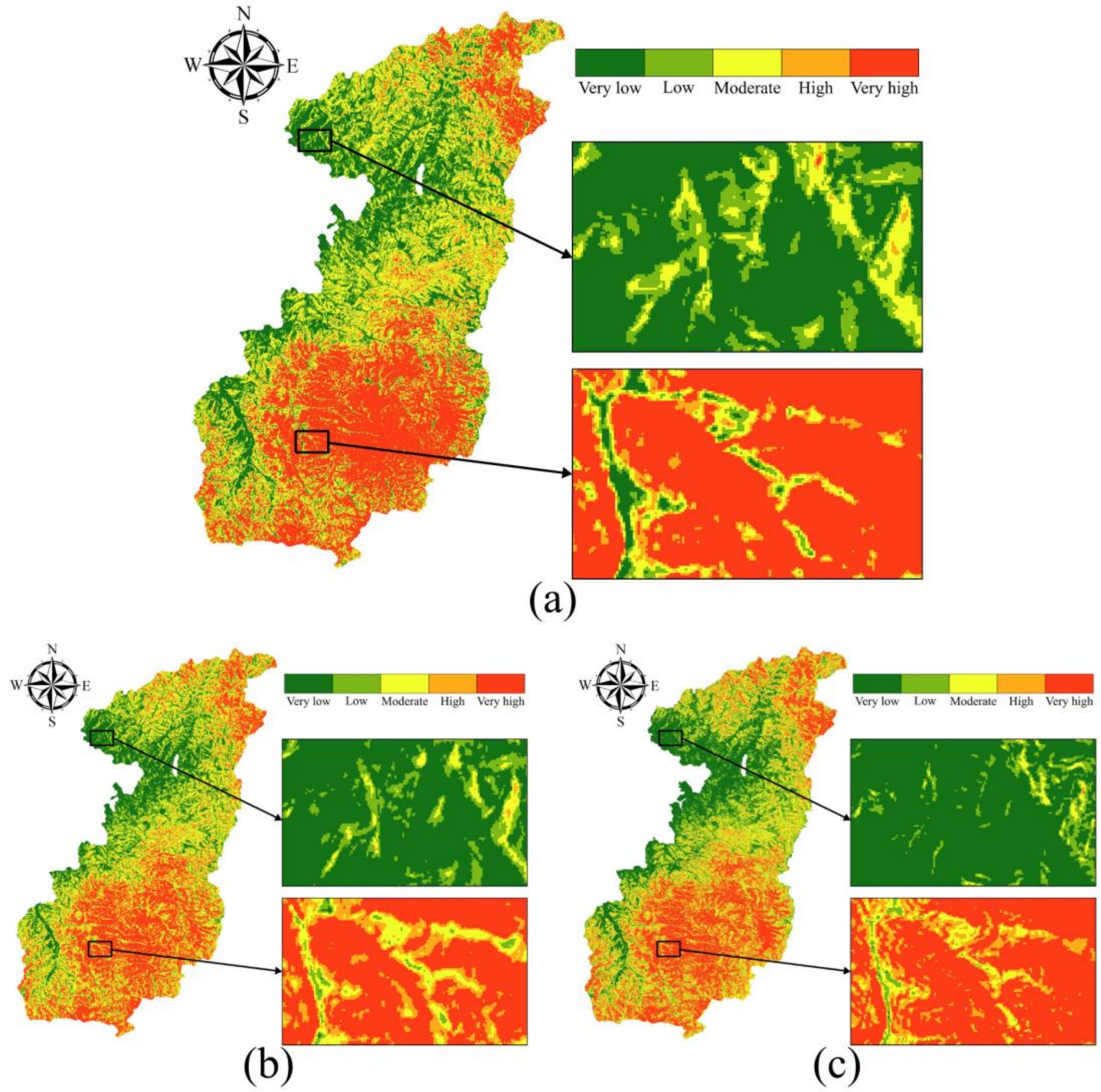

**Fig. 8**. Landslide susceptibility maps produced by (a) the LGSCF-based Model A+M, (b) Model A, (c) Model M.

## 4.4 Model performance analysis

### 4.4.1 Classification metrics and ROC-AUC results

The evaluation metrics of the best-performing configuration were averaged across its runs (Table 6). For ROC curves, which cannot be directly averaged across runs, the curve corresponding to the median AUC run was selected for plotting (Fig. 9).

Among the baseline models, the spatial-context branch networks (Models A, B, and C) consistently outperformed the local-geo branch networks (Models M, N, and P). In particular, Models B and C achieved the strongest performance, with accuracy and F1-scores exceeding 86% and AUC values above 0.945. By contrast, the pixel-based models, especially Models M and N, achieved relatively weaker overall performance. Although their precision values were relatively high (above 88%), their recall dropped below 78%, suggesting that these models tended to overemphasise one class and failed to provide balanced predictions across positive and negative samples.

When comparing the LGSCF-based models with their baselines, a clear and consistent improvement can be observed (Table 6). The F1-score and AUC are consistently higher for the LGSCF-based models than for their corresponding baselines, with the gains particularly evident for the weaker local-geo (Models M, N, and P). Even for the stronger spatial baselines (Models A, B, and C), the LGSCF-based models still achieved marginal but steady improvements. For example, Models B+N and C+P achieved AUC values of 0.9473 and 0.9472, respectively, both higher than their already strong baselines. This demonstrates that synergising local geo-environmental information with spatial context provides a complementary effect that enhances overall predictive performance. It not only compensates for weaker baselines but also further refines strong ones, highlighting the broad applicability of the LGSCF strategy. The ROC curves (Fig. 9) further confirm these findings, as the LGSCF-based models lifted the curves upward across most of the false positive rate range. Nevertheless, precision and recall still varied in some cases, indicating that the balance between detecting positives and negatives was not yet perfect. Even so, compared with the pixel-based baselines, the LGSCF-based models achieved a much more acceptable balance, approaching that of the spatial baselines. This also suggests that the characteristics of the two branch types exerted a potential influence on the performance of the LGSCF-based models.

**Table 6**. Averaged evaluation metrics of baseline and LGSCF-based models from the best-performing configuration (based on mean AUC value).

| Types | Models | Accuracy | Precision | Recall | F1-score | AUC |
|---|---|---|---|---|---|---|
| Spatial-context branch CNNs | A | 84.018% | 85.638% | 82.298% | 83.827% | 0.92043 |
| | B | 86.858% | 89.516% | 84.186% | 86.577% | 0.94506 |
| | C | 86.224% | 86.887% | 86.193% | 86.317% | 0.94537 |
| Local-geo branch CNNs | M | 83.633% | 87.969% | 78.354% | 82.805% | 0.91515 |
| | N | 83.452% | 88.129% | 77.741% | 82.537% | 0.91484 |
| | P | 83.657% | 90.514% | 75.518% | 82.297% | 0.92001 |
| LGSCF-based CNNs | A+M | 84.539% | 88.809% | 79.892% | 83.843% | 0.92716 |
| | A+N | 84.223% | 85.066% | 83.636% | 84.222% | 0.92510 |
| | A+P | 84.572% | 87.798% | 80.792% | 84.052% | 0.92621 |
| | B+M | 87.255% | 90.193% | 84.003% | 86.905% | 0.94720 |
| | B+N | 87.416% | 90.271% | 84.114% | 87.063% | 0.94640 |
| | B+P | 87.055% | 87.764% | 86.632% | 87.089% | 0.94662 |
| | C+M | 86.978% | 87.816% | 86.265% | 86.951% | 0.94604 |
| | C+N | 87.311% | 89.864% | 84.528% | 86.996% | 0.94668 |
| | C+P | 87.131% | 88.239% | 86.058% | 87.073% | 0.94694 |

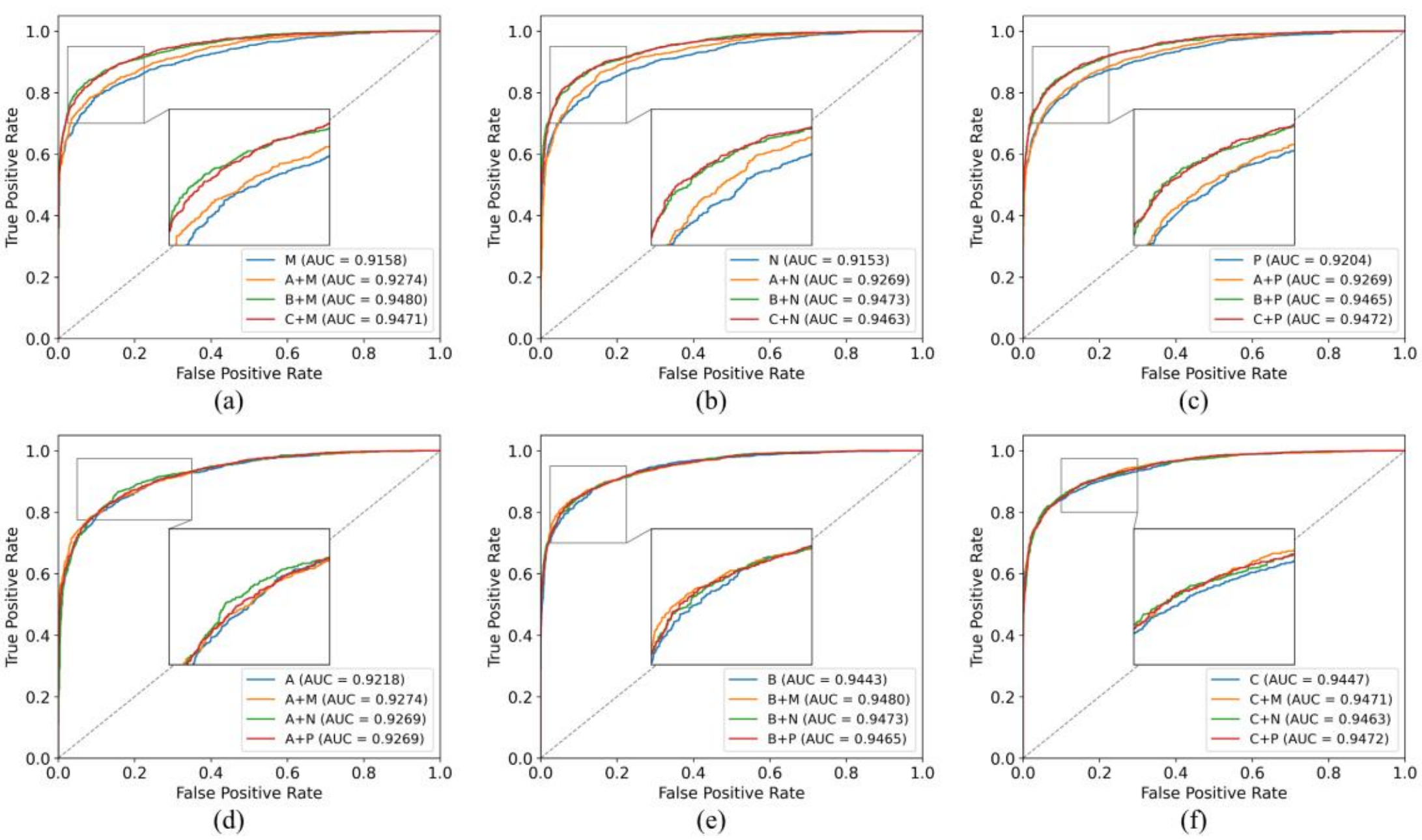

**Fig. 9**. ROC curves of baseline models and their LGSCF-based combinations: (a) Model M, (b) Model N, (c) Model P, (d) Model A, (e) Model B, (f) Model C. Further details of the base models and their LGSCF-based combinations are provided in Tables 3 and 4.

### 4.4.2 Error-based evaluation (RMSE and MAE)

Fig. 10 compares the RMSE and MAE of the baseline models and their corresponding LGSCF-enhanced variants. Overall, incorporating additional branches consistently reduced prediction errors relative to the single-branch baselines, indicating the effectiveness of the LGSCF strategy. Among the spatial-context branch models (Models A, B, and C), Model B achieved the most noticeable reductions in both RMSE and MAE, while improvements for Models A and C were more moderate. For the local-geo branch models (Models M, N, and P), the performance gains were particularly pronounced. This trend is consistent with the ROCï AUC results, indicating agreement between classification- and error-based evaluation strategies.

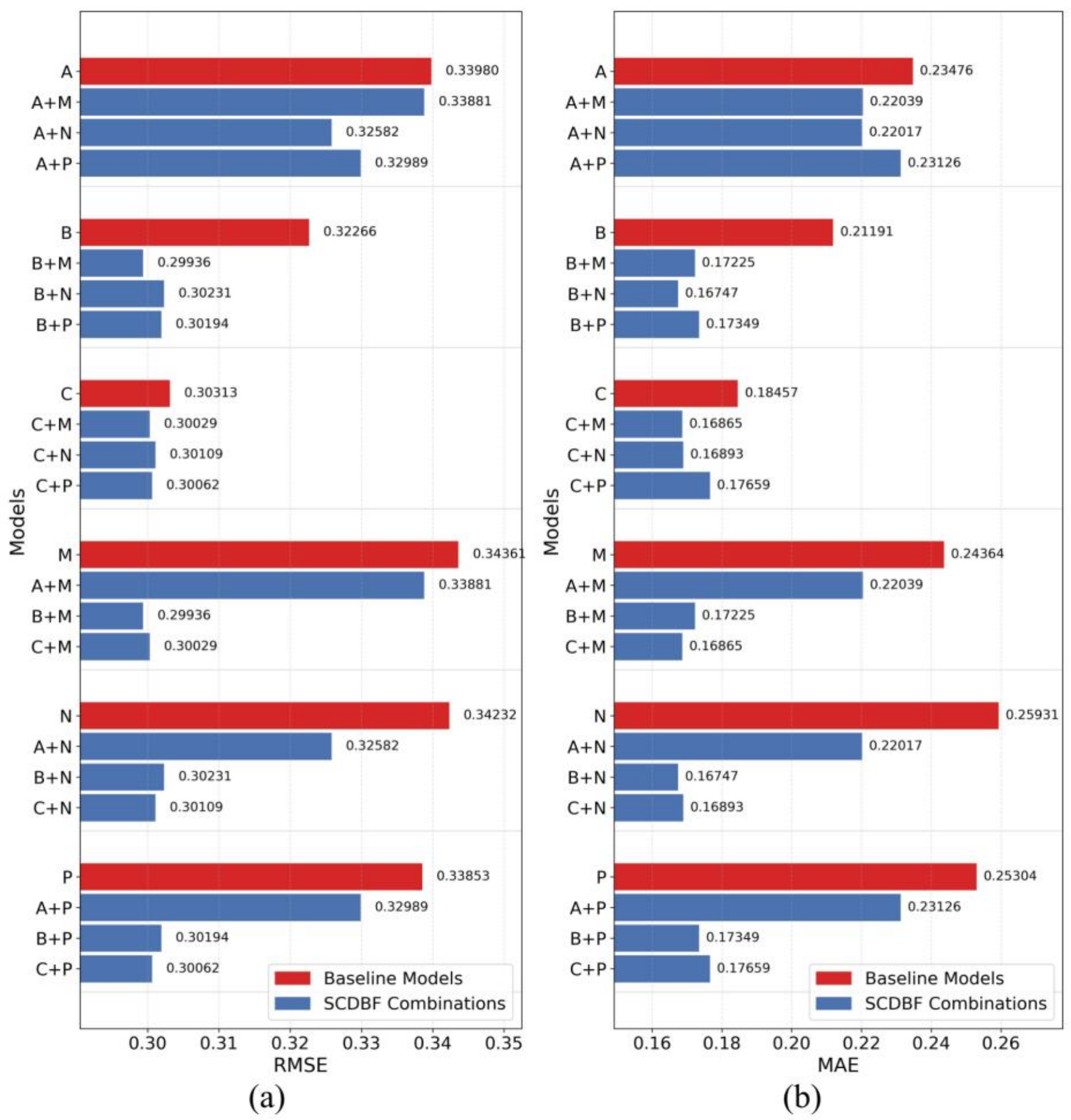

**Fig. 10**. RMSE and MAE of baseline models and their corresponding LGSCF-based combinations: (a) RMSE, (b) MAE.

## 4.5 Analysis of the Importance of Factors

Fig. 11 presents the SHAP interpretations of Models B, M, and B+M. In the beeswarm plots, each point represents an individual sample, while the feature-importance plots rank the LCFs according to their mean absolute SHAP values. Despite differences in their relative rankings, distance to faults, DEM, distance to rivers, slope, and rainfall consistently formed the dominant factor group across all three models. Model B placed greater emphasis on distance to faults, DEM, and distance to rivers (Fig. 11a,b), whereas Model M was driven more strongly by slope and rainfall (Fig. 11c,d). Following LGSCF fusion, Model B+M retained distance to faults as the leading factor while increasing the relative contributions of slope and rainfall (Fig. 11e,f). Thus, LGSCF did not introduce a different set of dominant factors but redistributed the contributions of the same core LCFs captured by the two branches. These factors collectively represent structural, topographic, and hydro-environmental controls, although their SHAP importance should not be interpreted as direct causality. DEM, for example, may act as a proxy for geomorphic position and the broader rainfallï erosion setting because it is associated with slope development, terrain dissection, orographic rainfall, and topographic energy. Similarly, distance to faults and rivers may represent the spatial influence of structural weakening and fluvial incision rather than isolated distance effects. Curvature-related factors, NDVI, geology, and TRI made comparatively limited contributions, suggesting that their predictive information was either secondary or partly represented by the dominant factors.

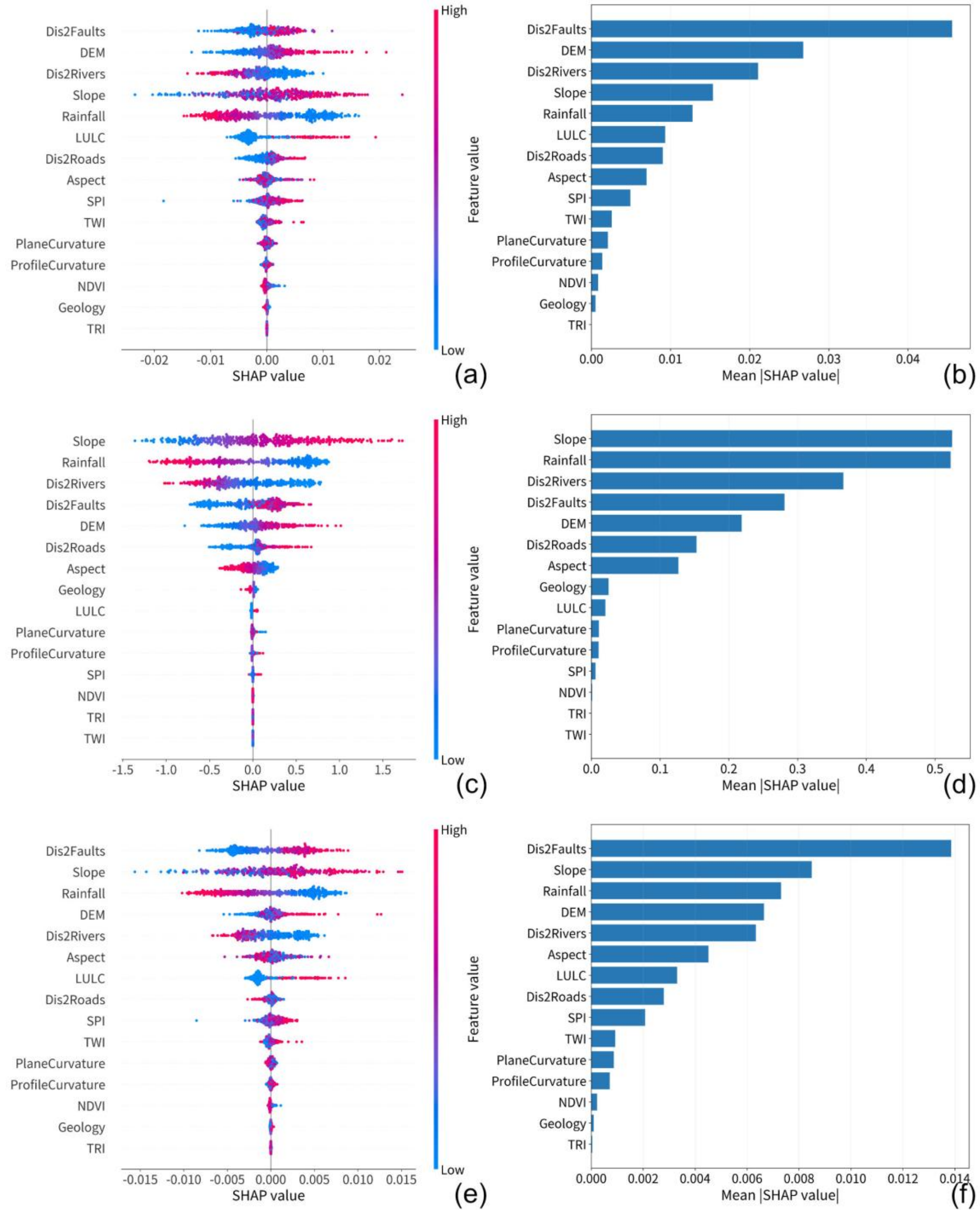

**Fig. 11**. SHAP-based interpretation of Models B, M, and B+M using beeswarm and mean absolute SHAP feature-importance plots: **(a)** and **(b)**: Model B; **(c)** and **(d)**: Model M; **(e)** and **(f)**: Model B+M.

# 5 Discussions

## 5.1 Comparison of landslide susceptibility maps

To further quantify how landslides are distributed across susceptibility levels, we calculated their percentages in each of the five classes (Table 7 and Fig. 12) and visualised the results for all baseline and LGSCF-based models. For the baseline models, around half of the landslides already fell in the "very high" susceptibility class, ranging from roughly 46% for Models M and N to over 55% for Models B and C. However, a considerable proportion was still scattered across the "moderate" and "high" classes, and some baselines allocated relatively large shares to the "low" and "very low" levels, particularly Models N, P, and B.

By contrast, the LGSCF-based models consistently increased the concentration of landslides in the "very high" class while reducing allocations to the lower susceptibility categories. For instance, Model A+M raised the proportion of "very high" susceptibility to 55.29%, compared with 52.76% and 45.91% for Models A and M, respectively. Similar improvements appeared across other combinations: Models B+M, B+N, and B+P all exceeded 59%, while Models C+N and C+P exceeded 62%. Model C+N reached the highest percentage of 66.84% in the "very high" susceptibility class, marking a clear improvement over both of its baselines. At the same time, the LGSCF-based models substantially reduced misclassifications in the lower categories. Whereas baseline models placed about 2% of landslides in the "very low" susceptibility class and, in some cases more than 10% in the "low" susceptibility class, the LGSCF-based models reduced the "very low" share to around 1%, while the combined total of "very low" and "low" is as small as roughly 5%.

**Table 7**. Percentage of landslides within each susceptibility class.

| Models | Very low | Low | Moderate | High | Very high |
|---|---|---|---|---|---|
| Model A | 1.44% | 5.25% | 19.49% | 21.06% | 52.76% |
| Model B | 1.74% | 14.10% | 12.60% | 15.92% | 55.63% |
| Model C | 2.44% | 5.38% | 20.35% | 16.56% | 55.27% |
| Model M | 2.78% | 9.49% | 19.04% | 22.79% | 45.91% |
| Model N | 3.56% | 16.75% | 15.98% | 17.65% | 46.06% |
| Model P | 2.72% | 10.17% | 16.11% | 16.79% | 54.22% |
| Model A+M | 1.43% | 4.89% | 21.94% | 16.45% | 55.29% |
| Model A+N | 0.90% | 5.14% | 20.39% | 14.72% | 58.85% |
| Model A+P | 1.24% | 5.12% | 20.56% | 15.17% | 57.91% |
| Model B+M | 0.54% | 3.71% | 17.12% | 13.99% | 64.63% |
| Model B+N | 0.81% | 2.78% | 19.56% | 17.07% | 59.79% |
| Model B+P | 0.88% | 3.30% | 17.87% | 13.99% | 63.95% |
| Model C+M | 1.09% | 2.78% | 18.14% | 13.63% | 64.37% |

| Model C+N | 0.88% | 2.81% | 16.20% | 13.26% | 66.84% |
|---|---|---|---|---|---|
| Model C+P | 1.09% | 3.17% | 18.19% | 15.40% | 62.15% |

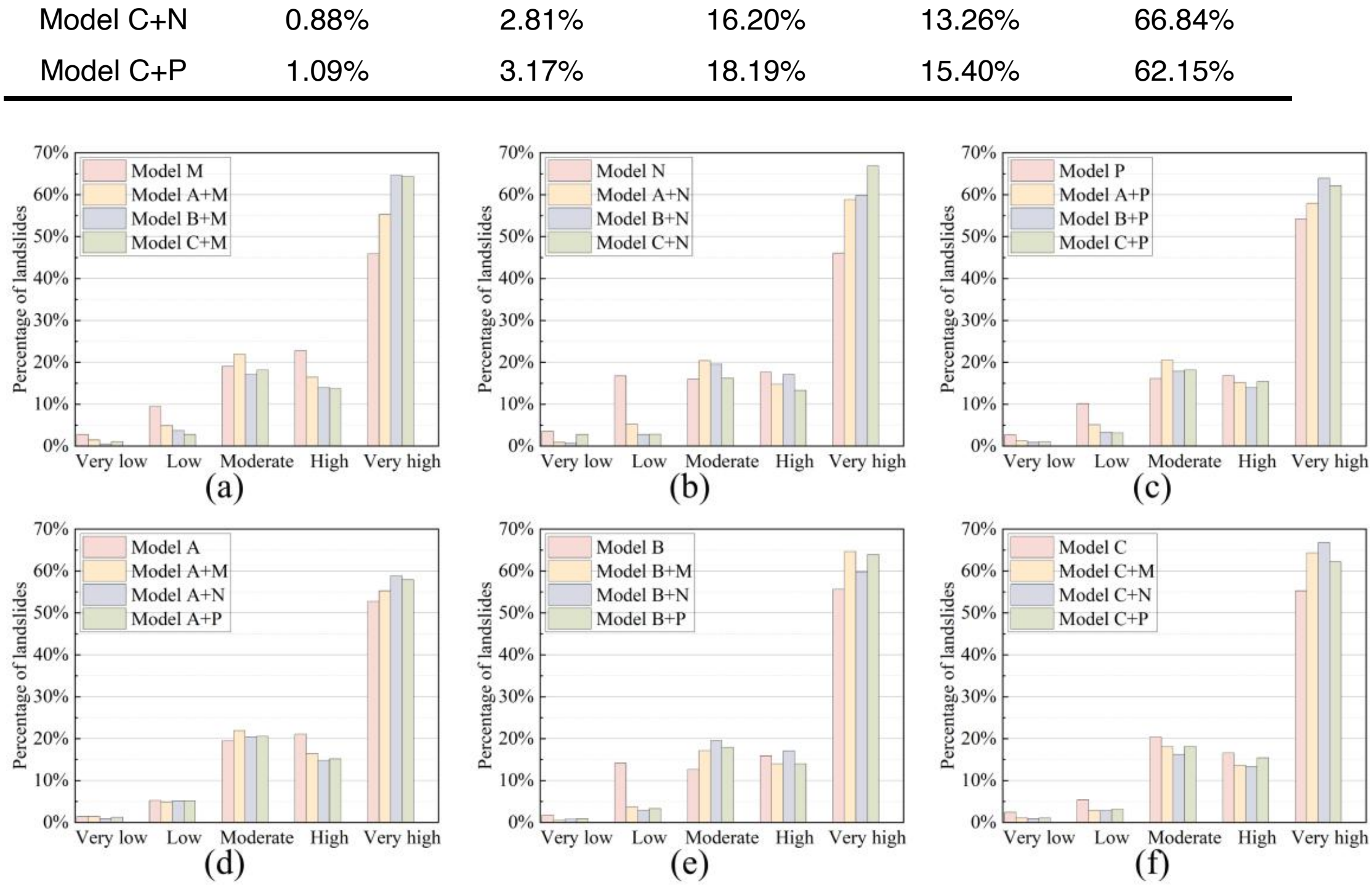

**Fig. 12**. Percentage of landslides within each susceptibility class for the baseline and corresponding LGSCF-based models: (a) Model M and its LGSCF combinations, (b) Model N and its LGSCF combinations, (c) Model P and its LGSCF combinations, (d) Model A and its LGSCF combinations, (e) Model B and its LGSCF combinations, (f) Model C and its LGSCF combinations. The LGSCF-based models generally show higher proportions of landslides in the "very high" class and lower proportions in the "very low" class. Further details of the baseline models and their corresponding LGSCF combinations are provided in Tables 3 and 4.

## 5.2 Robustness and additional regional evaluation

To further assess the robustness and spatial generalisability of the results, a five-fold spatial robustness analysis was conducted using the original experimental fold and four additional spatially partitioned folds. Identical fold configurations were applied to all models. Fold-wise AUC values were summarised using the mean, standard deviation, and t-based 95% confidence interval. As each LGSCF-based model and its corresponding spatial-context baselines (Model A, B, or C) were evaluated using the same folds, paired t-tests were used to assess whether the observed AUC improvements were statistically significant.

The five-fold results were broadly consistent with the original random-split evaluation (Fig. 13), and Model B+M remained the best-performing combination, achieving a mean AUC of 0.9506. All nine LGSCF-based models achieved higher mean AUC values than their corresponding spatial-context baselines. Eight of the nine LGSCF-based models exhibited lower standard deviations (SD) than their corresponding spatial-context baselines, indicating that their AUC values were generally less sensitive to changes in the spatial data partition. Seven combinations showed statistically significant improvements at $p < 0.05$, the gains for Model B+P (p = 0.1699) and Model C+N (p = 0.1233) were not significant, indicating that the benefit of LGSCF is not equally strong for every branch pairing.

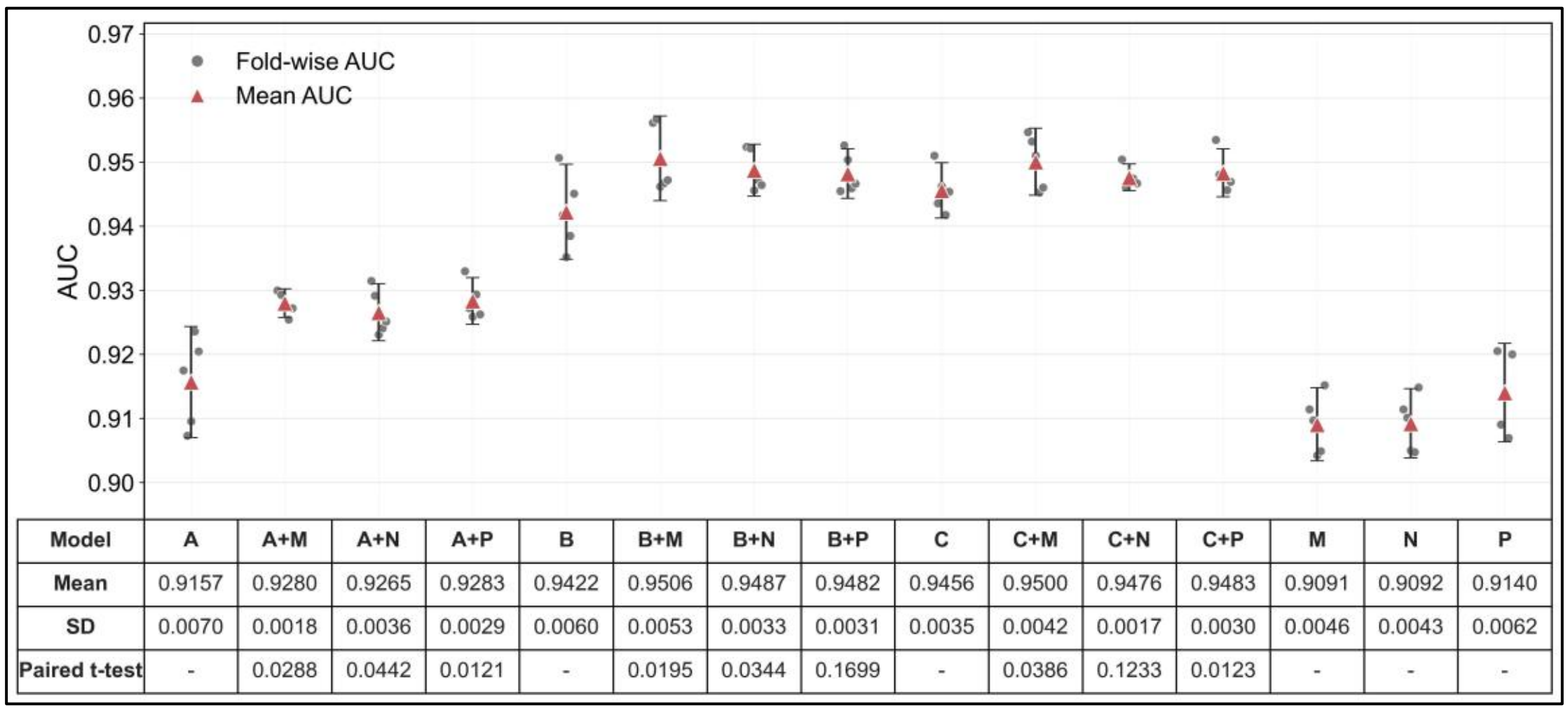

| Model | A | A+M | A+N | A+P | B | B+M | B+N | B+P | C | C+M | C+N | C+P | M | N | P |
|---|---|---|---|---|---|---|---|---|---|---|---|---|---|---|---|
| Mean | 0.9157 | 0.9280 | 0.9265 | 0.9283 | 0.9422 | 0.9506 | 0.9487 | 0.9482 | 0.9456 | 0.9500 | 0.9476 | 0.9483 | 0.9091 | 0.9092 | 0.9140 |
| SD | 0.0070 | 0.0018 | 0.0036 | 0.0029 | 0.0060 | 0.0053 | 0.0033 | 0.0031 | 0.0035 | 0.0042 | 0.0017 | 0.0030 | 0.0046 | 0.0043 | 0.0062 |
| Paired t-test | - | 0.0288 | 0.0442 | 0.0121 | - | 0.0195 | 0.0344 | 0.1699 | - | 0.0386 | 0.1233 | 0.0123 | - | - | - |

**Fig. 13.** AUC values under 5-fold spatial cross-validation. Error bars indicate the 95% confidence intervals. The table reports the mean AUC, standard deviation, and paired t-test p-values. Paired t-tests were conducted with the corresponding baseline model as the reference: Model A for Models A+M, A+N, and A+P; Model B for Models B+M, B+N, and B+P; and Model C for Models C+M, C+N, and C+P. Therefore, p-values are reported only for the LGSCF-based models.

In addition, to assess whether the effectiveness of the LGSCF framework could be reproduced beyond the primary study area, an evaluation was conducted at an additional study site located in Hualien County. The Hualien experiment followed the same modelling procedure as that used for the primary study area, including consistent data sources and the same set of 15 LCFs, as well as identical strategies for non-landslide sampling and dataset construction. Although the overall predictive performance in Hualien was lower than that obtained in the primary study area, the relative advantage of LGSCF was largely preserved (Fig. 14). All nine LGSCF-based models achieved higher AUC values than their corresponding baselines. F1-scores also increased for eight of the nine combinations, with only Model C+N showing a slight decrease of 0.06 percentage points. These results indicate that the performance gains associated with integrating local geo-environmental characteristics and spatial context were not confined to the Nantou study area. Together with the cross-validation, this additional evaluation provides further evidence for the robustness and broader applicability of the LGSCF framework, although the magnitude and statistical significance of its benefits varied among different combinations.

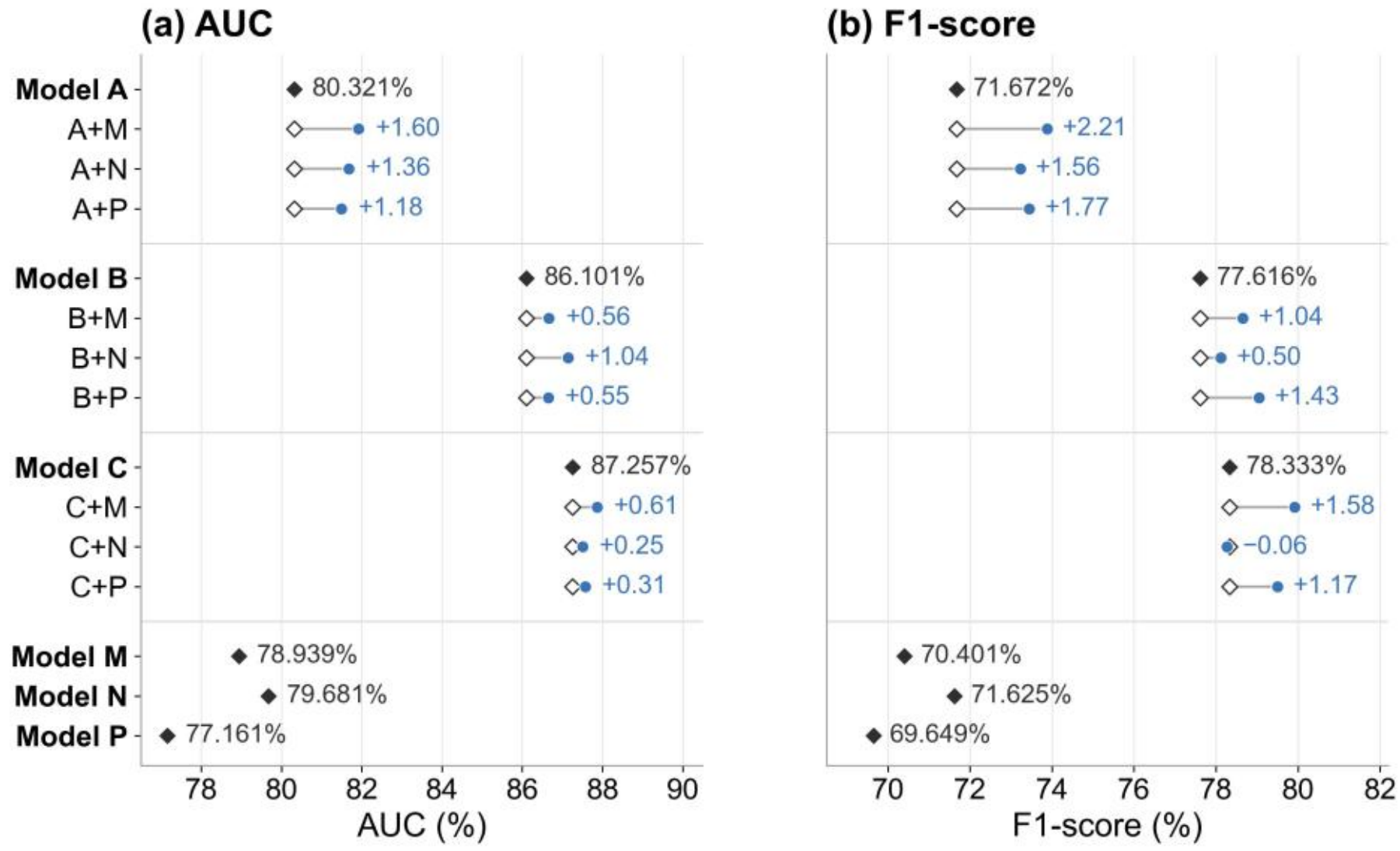

**Fig. 14**. Evaluation of the LGSCF framework in the additional study area of Hualien County in terms of (a) AUC and (b) F1-score.

## 5.3 Mechanism analysis

As demonstrated in the results section, the LGSCF strategy consistently improved model performance and enhanced the LSM task. This gain can be largely attributed to the complementary information captured by the spatial-context and local-geo branches. The spatial-context branch learns neighbourhood morphology, slope continuity, and geomorphic context, while the local-geo branch focuses on point-level features that are strongly associated with landslide initiation. LGSCF enables pixel-level cues to adaptively modulate spatial features, allowing the model to selectively emphasise informative regional patterns when local instability signals are present. In this way, the interaction supplements spatial context and helps mitigate cases where spatial features alone may be ambiguous or not directly relevant to landslide occurrence.

To further clarify the mechanism of LGSCF, Table 8 compares several approaches according to how local-geo and spatial context information are modelled and integrated. Pixel-based CNN1D models explicitly extract the LCF characteristics of the target location but do not incorporate neighbourhood information, whereas patch-based CNNs capture spatial context while implicitly mixing the target location with surrounding features. Attention-enhanced CNNs further recalibrate contextual representations, but the local and contextual information remain within the same feature stream. Dual-branch approaches preserve the two representations separately. However, direct concatenation does not explicitly account for the relationship between the two feature representations. The spatial-context branch already incorporates information from the target pixel, while the two branches encode information at different spatial scopes. Simply stacking their features leaves both the overlapping information and the cross-branch relationship to be resolved implicitly by subsequent layers. In contrast, LGSCF establishes an explicit directional interaction, using local-geo features to modulate the spatial-context representation through feature-wise scaling and shifting.

**Table 8.** Comparison of representative local-geo, spatial-context, and feature-integration strategies in LSM.

| Method | Input | Local information modelling | Spatial-context modelling | Local-geo and spatial context integration |
|---|---|---|---|---|
| Pixel-based CNN1D | Target pixel LCF vector | Explicit point-level feature extraction | Not included | None |
| Patch-based CNN (CNN2D and CNN3D) | LCF patch | Implicit; target-location information is embedded in the patch | Convolutional extraction | Implicit mixing during convolution and pooling |
| Attention-enhanced CNN (e.g., CBAM) | LCF patch | Implicit within the contextual representation | Convolution with channel/spatial feature recalibration | Within-stream feature recalibration |
| Dual-branch concatenation | Target pixel LCF vector + LCF patch | Separate local feature extraction | Separate spatial-context feature extraction | Direct concatenation of local and contextual features |
| Proposed LGSCF | Target pixel LCF vector + LCF patch | Separate local feature extraction | Separate spatial-context feature extraction | FiLM-based directional modulation |

To examine whether this conceptual distinction is reflected in model performance, an additional experiment was conducted using Model A as the spatial-context branch and Models M, N, and P as the respective local-geo branches, with direct concatenation replacing the proposed FiLM-based fusion. The corresponding LGSCF results were taken from the runs shown in Fig. 9 for comparison. As shown in Fig. 15, the three concatenation-based combinations achieved AUC values comparable to those of the corresponding local-geo branches but did not outperform the spatial-context baseline (Model A). A similar pattern was observed for F1-score, for which concatenation provided only limited improvement, with only Concatenation A+M exceeding Model A and the corresponding LGSCF variant. These results suggest that simply increasing the amount of feature information does not necessarily lead to an effective fusion and may provide little or even adverse benefit relative to the stronger constituent branch. Moreover, the varying gains among different LGSCF combinations indicate that fusion performance is not determined solely by the predictive strength of the individual branches, but also by how effectively their complementary representations are coordinated.

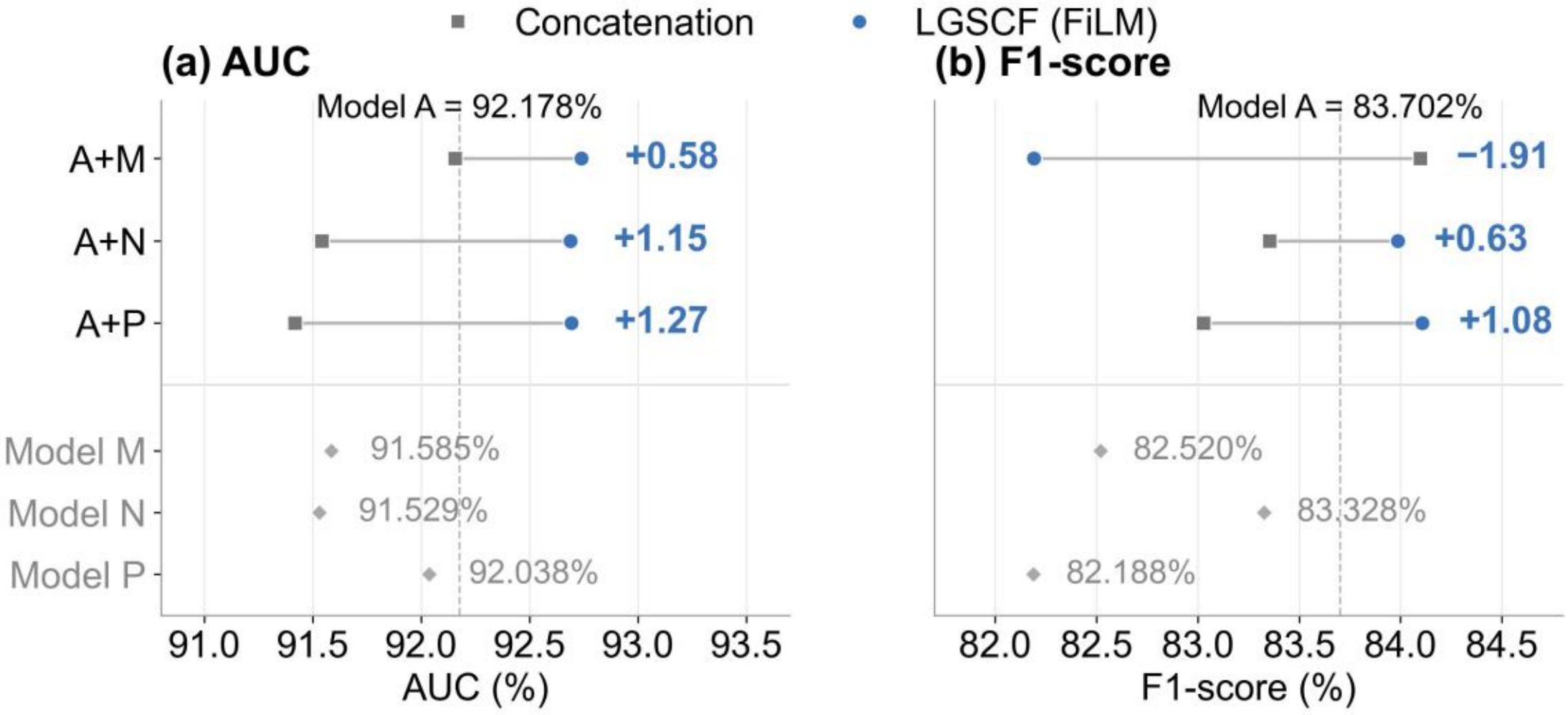

**Fig. 15**. Comparison of direct concatenation and FiLM-based LGSCF in terms of (a) AUC and (b) F1-score.

## 5.4 Limitations and future work

Several limitations of the LGSCF framework should be acknowledged. First, although the performance gains using the LGSCF strategy were clearly observed, the contributions of individual FiLM components were not quantitatively isolated. Future work could therefore conduct finer-grained ablation experiments, including scaling-only and shifting-only configurations, to further clarify the respective roles of these components. Second, for baseline models with already strong performance (e.g., Model C), the improvement brought by LGSCF becomes relatively modest. This raised a potential need to investigate how the strategy behaves when using more advanced architectures as one of the branches, and whether additional refinement or adaptive mechanisms may be required under such settings. Third, while LGSCF generally enhanced overall prediction capability, the improvement across positive and negative classes was not always symmetric. This imbalance might stem from characteristics inherited from the baselines, suggesting that future work could explore strategies to further enhance class-balanced prediction performance. Finally, the framework was evaluated in a single study area. Finally, although the additional evaluation in Hualien County showed that the performance gains of LGSCF could be reproduced beyond the primary study area, both study areas are located in Taiwan and therefore represent a relatively limited geographical context. Further evaluation in regions with more distinct geological, climatic, and landslide conditions is still required to establish the broader applicability of the framework.

In addition, LGSCF is subject to several sources of uncertainty commonly encountered in LSM. For instance, the 11 Ĭ 11 patch size was selected according to the typical landslide extent within the study area and the spatial resolution of the LCFs. However, different patch sizes may change the amount of neighbourhood information captured by the model and consequently affect its performance, which requires further investigation [83]. Non-landslide samples were selected using the FR method. Although this approach reduces the likelihood of incorrectly labelling potential landslide locations as stable terrain, it may favour negative samples that are environmentally distinct from landslide locations. Various alternative methods for selecting non-landslide samples have been investigated in previous studies [68,69,84ï 86], but which method is most suitable for the proposed strategy remains unclear. The landslide inventory also introduces potential uncertainty through omissions, positional inaccuracies, temporal mismatches with the conditioning-factor data, and the simplification involved in representing landslide areas using selected sample locations [55,87]. Future work should therefore examine the sensitivity of LGSCF to different patch sizes and negative-sample selection strategies and evaluate the framework using multi-temporal and field-validated landslide inventories.

## 6 Conclusion

This study proposed the LGSCF strategy, which integrates dual-branch CNNs through a feature-wise modulation mechanism to jointly capture local-geo characteristics of landslide or non-landslide locations and neighbourhood spatial context in LSM. Instead of defining a single deep learning network structure, LGSCF provides a flexible framework that can be applied to integrate different CNN architectures, allowing for broad adaptability.

Six baseline models were used to validate the effectiveness of the LGSCF strategy. The experimental results show consistent improvements in LSM accuracy using the LGSCF-based models over their single-branch counterparts. In particular, the LGSCF-based models achieved higher F1-scores and AUC values, and showed a more pronounced concentration of landslides in the very high susceptibility category. The cross-validation and additional evaluation in Hualien County demonstrated the robustness and reproducibility of the LGSCF strategy. These findings demonstrate that synergising local geo-environmental characteristics with spatial context yielded more accurate predictions. From a practical perspective, these outcomes suggest that existing CNN-based LSM models can be readily enhanced by incorporating the LGSCF strategy, offering improved predictive accuracy. Nevertheless, although the additional Hualien evaluation extends the assessment beyond the primary study area, both study areas are located in Taiwan, and further evaluation under more diverse geological, climatic, and landslide conditions remains necessary to assess the broader applicability of the proposed framework. It is also valuable to systematically examine which data characteristics, model structures, or parameter settings most strongly influence the effectiveness of the proposed strategy. Such efforts can help refine the framework and maximise its benefits.

## Author contributions

**Y.C.**: Conceptualisation, methodology, software, formal analysis, data curation, investigation, visualisation, writing - original draft, writingð review and editing. **L.F.**: Conceptualisation, methodology, software, formal analysis, data curation, investigation, supervision, writingð review and editing. **Q.Z.**: Conceptualisation, methodology, formal analysis, data curation, software, investigation, writingð review and editing. **C.Z.**: Conceptualisation, methodology, formal analysis, investigation, writingð review and editing. **Y.L.**: Conceptualisation, writingð review and editing. **R.M.**: Conceptualisation, writingð review and editing.

## Funding

This research was supported by the Xiâan Jiaotong-Liverpool University Postgraduate Research Scholarship under Grant FOSA2312049.

## Data availability

The datasets used and/or analysed during the current study will be made available from the corresponding author on reasonable request.

## Declarations

### Ethical approval

Not applicable.

### Consent to participate

Not applicable.

### Consent to publications

Not applicable.

### Declaration of competing interest

The authors declare that they have no competing interests or other interests that might be perceived to influence the results and/or discussion reported in this paper.

# Appendix A

(a) (b)

(c) (d)

(e) (f)

**Fig. A1**. Landslide susceptibility maps produced by six base models: (a) Model A, (b) Model B, (c) Model C, (d) Model M, (e) Model N, and (f) Model P.

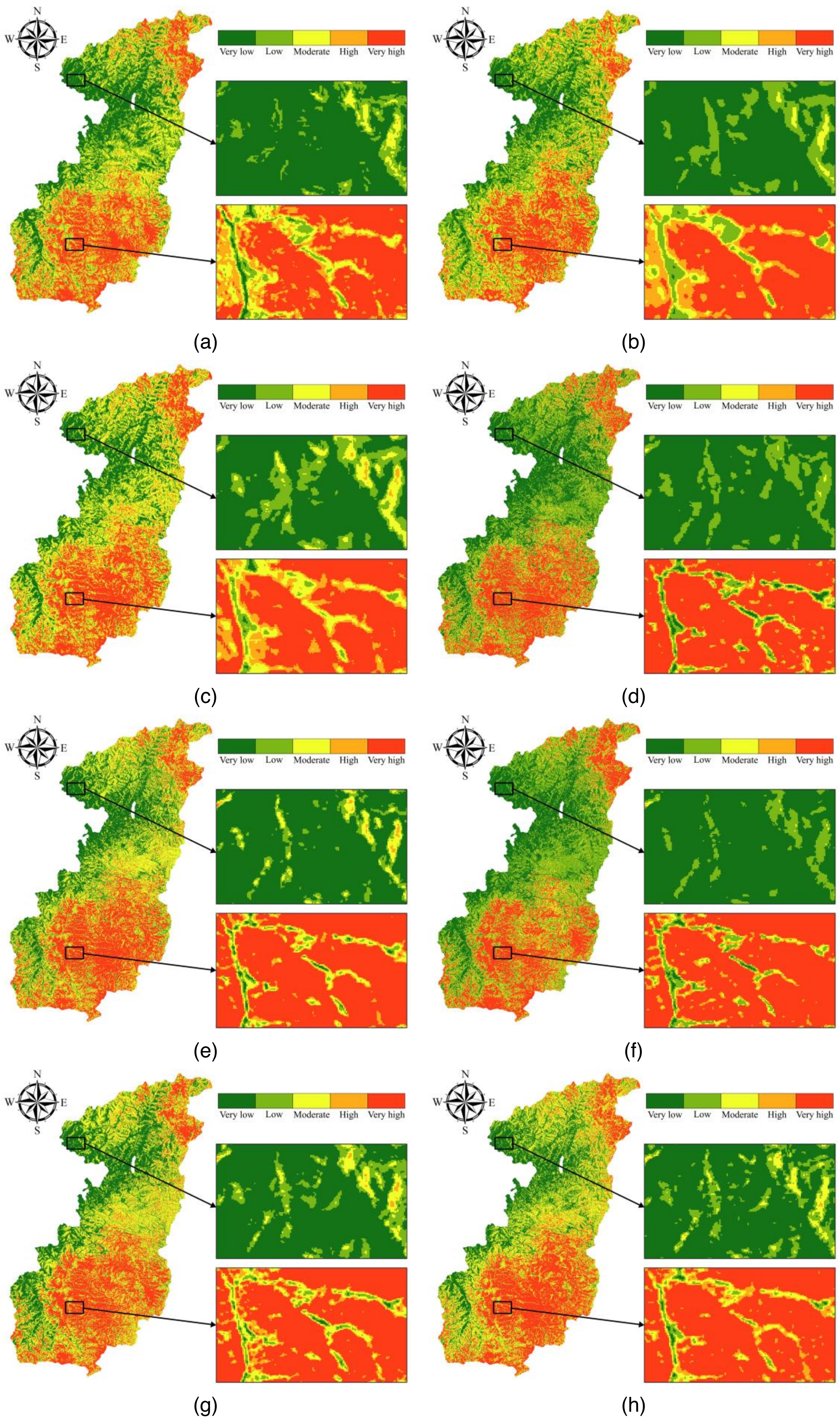
N
W
E
S
Very low
Low
Moderate
High
Very high
(a)
(b)
(c)
(d)
(e)
(f)
(g)
(h)

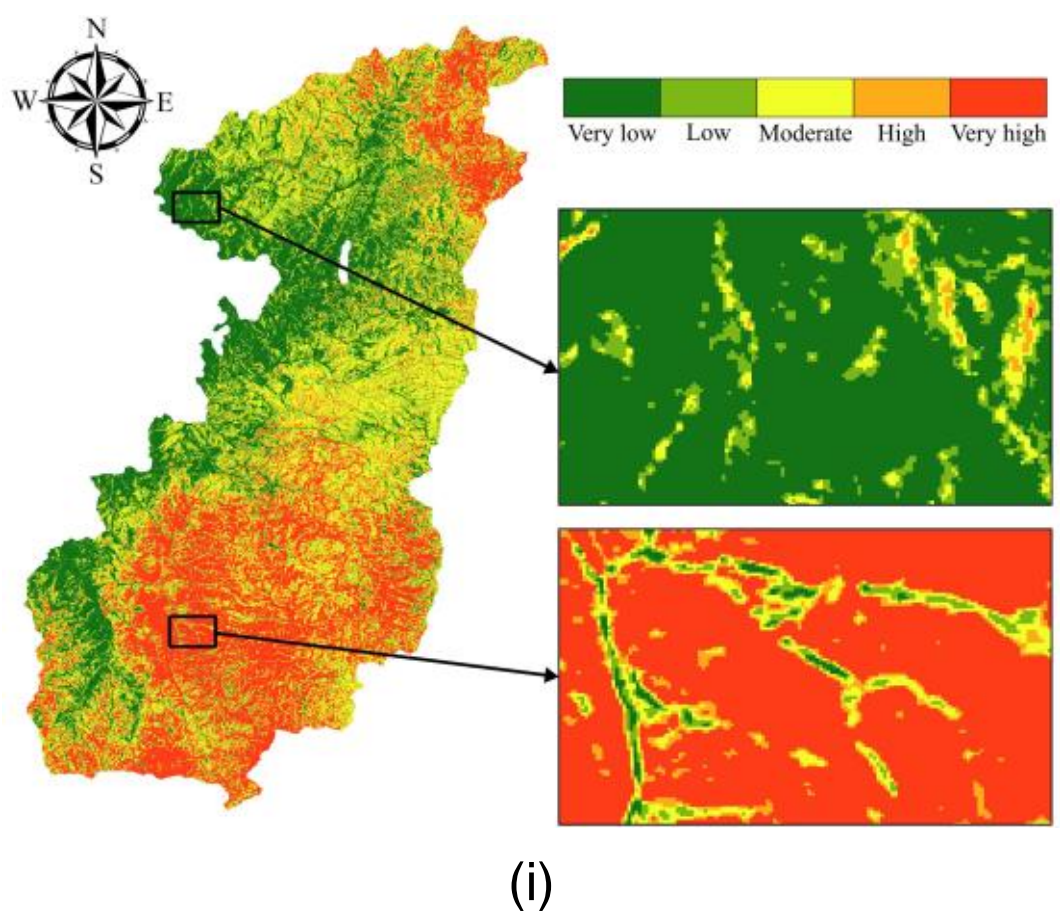

(i)

**Fig. A2**. Landslide susceptibility maps produced by nine LGSCF-based models: (a) Model A+M, (b) Model A+N, (c) Model A+P, (d) Model B+M, (e) Model B+N, (f) Model B+P, (g) Model C+M, (h) Model C+N, and (i) Model C+P.

## Appendix B

**Table B1**. Hardware configuration, training time, and computational complexity of the baseline and LGSCF models.

| Items | M | N | P | A | B | C | A+M | A+N | A+P | B+M | B+N | B+P | C+M | C+N | C+P |
|---|---|---|---|---|---|---|---|---|---|---|---|---|---|---|---|
| Time (s) | 177.58 | 287.08 | 161.69 | 291.36 | 187.69 | 177.82 | 300.93 | 232.91 | 240.40 | 170.31 | 148.69 | 132.01 | 143.53 | 140.49 | 145.42 |
| Parameters | 6252 | 6514 | 1113890 | 150596 | 77980 | 39618 | 197637 | 214602 | 490234 | 93000 | 129336 | 249876 | 80742 | 112210 | 238946 |
| FLOPs | 14882 | 20642 | 5862818 | 1536830 | 2209676 | 2024130 | 2549015 | 2861855 | 14847306 | 2241696 | 2319604 | 10568108 | 2108310 | 2176482 | 10377378 |
| CPU | AMD EPYC 7282 16-Core Processor | | | | | | | | | | | | | | |
| GPU | NVIDIA GeForce RTX 4090 D (24 GB VRAM) | | | | | | | | | | | | | | |
| Memory | 128 GB | | | | | | | | | | | | | | |
| Hard disk | 2 TB | | | | | | | | | | | | | | |

**Notes**: An early-stopping strategy with a patience of 10 epochs was applied during training. Specifically, training was terminated when the monitored validation performance did not improve for 10 consecutive epochs. Therefore, the reported training time represents the total time required for each model to reach its stopping condition, rather than the time required to complete a fixed number of epochs. The generally higher computational complexity of the LGSCF-based models is primarily attributable to their dual-branch architecture. Unlike the single-branch baselines, each LGSCF-based model simultaneously employs the feature-extraction components of its two corresponding baseline models. One branch preserves the characteristics of the pixel directly associated with the sample label, whereas the other extracts complementary spatial context from the surrounding neighbourhood. The features generated by the two branches are subsequently integrated through additional feature fusion operations. Such parallel feature extraction and subsequent fusion inevitably require additional computational operations and, in most cases, introduce additional trainable parameters. Despite this increase in theoretical complexity, the observed training times of the LGSCF models remained within a range comparable to those of the corresponding baseline models, without imposing a substantial additional training burden. This indicates that the integration of local-geo characteristics and neighbourhood context can be achieved at an acceptable computational cost, supporting the computational practicality and architectural rationality of the proposed LGSCF strategy.